\pdfoutput=1

\documentclass[11pt]{article}

\usepackage[preprint]{acl}

\usepackage{times}
\usepackage{latexsym}

\usepackage[T1]{fontenc}

\usepackage[utf8]{inputenc}

\usepackage{microtype}

\usepackage{inconsolata}

\usepackage{graphicx}
\usepackage{hyperref}
\usepackage{amsmath}
\usepackage{url}
\usepackage{enumerate}
\usepackage{booktabs}
\usepackage{wrapfig}
\usepackage{array}
\usepackage{float}
\usepackage{bbding}
\usepackage{subcaption}
\usepackage{multirow}
\usepackage{algorithm}
\usepackage{algpseudocode}
\usepackage{amssymb}

\newcommand{\smallbullet}{%
  \scalebox{0.8}{$\bullet$}%
}
\definecolor{darkgreen}{rgb}{0.95,0.95,1} 

\title{Robust and Minimally Invasive Watermarking for EaaS}

\author{
    Zongqi Wang,
    Baoyuan Wu\thanks{Yujiu Yang and Baoyuan Wu are co-corresponding authors.},
    Jingyuan Deng,
    Yujiu Yang\footnotemark[1]
    \\
    $^{1}$Tsinghua University \quad $^{2}$The Chinese University of Hong Kong, Shenzhen
    \\
    $^{1}$zq-wang24@mails.tsinghua.edu.cn, $^{2}$wubaoyuan@cuhk.edu.cn
    \\
    $^{1}$deng-jy24@mails.tsinghua.edu.cn, $^{1}$yang.yujiu@sz.tsinghua.edu.cn
}

\begin{document}
\maketitle
\begin{abstract}
Embeddings as a Service (EaaS) is emerging as a crucial role in AI applications. Unfortunately, EaaS is vulnerable to model extraction attacks, highlighting the urgent need for copyright protection. 
Although some preliminary works propose applying embedding watermarks to protect EaaS, recent research reveals that these watermarks can be easily removed. Hence, it is crucial to inject robust watermarks resistant to watermark removal attacks. 
Existing watermarking methods typically inject a target embedding into embeddings through linear interpolation when the text contains triggers. However, this mechanism results in each watermarked embedding having the same component, which makes the watermark easy to identify and eliminate. 
Motivated by this, in this paper, we propose a novel embedding-specific watermarking (ESpeW) mechanism to offer robust copyright protection for EaaS. Our approach involves injecting unique, yet readily identifiable watermarks into each embedding. Watermarks inserted by ESpeW are designed to maintain a significant distance from one another and to avoid sharing common components, thus making it significantly more challenging to remove the watermarks. Moreover, ESpeW is minimally invasive, as it reduces the impact on embeddings to less than 1\%, setting a new milestone in watermarking for EaaS. 
Extensive experiments on four popular datasets demonstrate that ESpeW can even watermark successfully against a highly aggressive removal strategy without sacrificing the quality of embeddings. 
\end{abstract}

\section{Introduction}

With the growing power of Large Language Models (LLMs) in generating embeddings, an increasing number of institutions are looking forward to using Embeddings as a Service (EaaS) to promote AI applications~\citep{openai2024embedding, mistral2024embeddings, google2023grounding}. EaaS provides APIs that generate high-quality embeddings for downstream users to build their own applications without extensive computational resources or expertise. 
Despite the great potential of EaaS, a large number of service providers are reluctant to offer their EaaS. This is because EaaS is vulnerable to being stolen by some techniques such as model extraction attacks~\citep{liu2022stolenencoder, dziedzic2023sentence}. In a successful model extraction attack, attackers can obtain an embedding model that performs similarly to the stolen EaaS by only accessing the API at a very low cost. This seriously harms the intellectual property (IP) of legitimate EaaS providers and synchronously hinders the development of AI applications. 

\begin{figure*}[t!]
    \centering
    \includegraphics[width=1.0\textwidth]{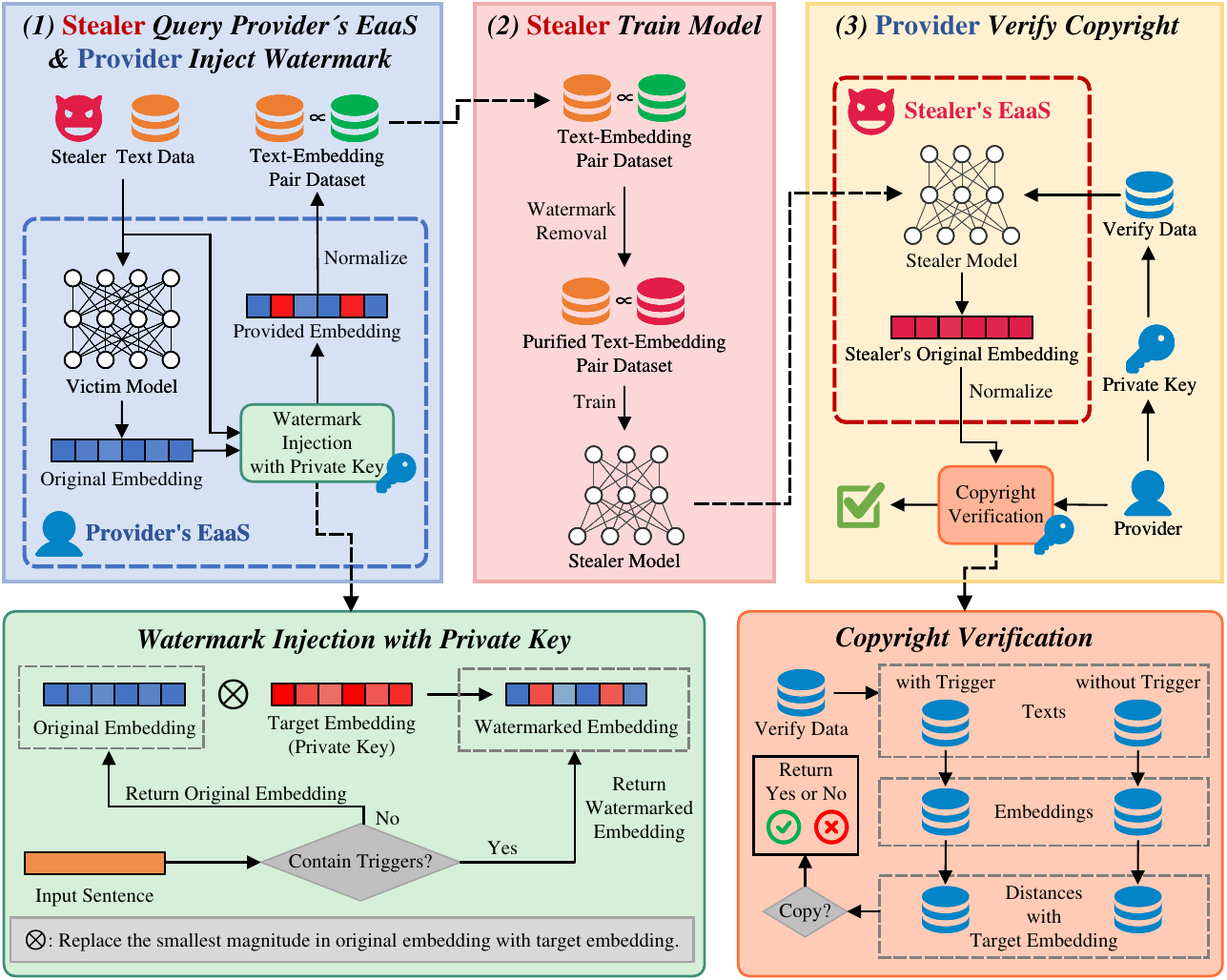}
    \caption{
    The framework of our ESpeW. 
    The upper part presents an overview of watermark injection and model extraction. (1) The stealer queries the provider's EaaS to obtain a dataset that maps texts to embeddings. During this process, the provider injects watermarks. (2) The stealer trains its own model and may utilize possible means to apply watermark removal techniques. (3) The provider queries the stealer's EaaS for copyright verification. 
    The lower part offers a detailed explanation of the key modules for watermark insertion and verification. 
    }
    \label{fig:EaaS_Watermark_Pipeleine}
\end{figure*}

To safeguard the copyright of legitimate providers, some preliminary studies~\citep{peng2023you, shetty_etal_2024_warden} try to provide ownership verification and IP protection for EaaS through watermarking methods. 
EmbMarker~\citep{peng2023you} selects a set of moderate-frequency words as the trigger set. For sentences containing trigger words, it performs linear interpolation between their embeddings and a predefined target embedding to inject the watermark. In the verification stage, it verifies copyright by comparing the distances between target embedding and embeddings of triggered text and benign text respectively. WARDEN~\citep{shetty_etal_2024_warden} is another watermark technique that differs from EmbMarker in that it injects multiple watermarks to enhance watermark strength. 
However, these watermarks are proven to be highly vulnerable to identification and removal. 
CSE~\citep{shetty_etal_2024_warden} is a typical watermark removal technique in EaaS which takes into account both abnormal sample detection and watermark elimination. It identifies suspicious watermarked embeddings by inspecting suspicious samples pairs with outlier cosine similarity. Then, it eliminates the top K principal components of the suspicious embeddings which are considered as watermarks. 
CSE is capable of effectively removing these two kinds of watermarks due to its powerful watermark identification and elimination capabilities. Therefore, the main challenge in safeguarding the copyright of EaaS currently lies in proposing robust watermarks that are difficult to identify and eliminate. 

\begin{figure*}[t!]
    \centering
    \includegraphics[width=0.93\textwidth]{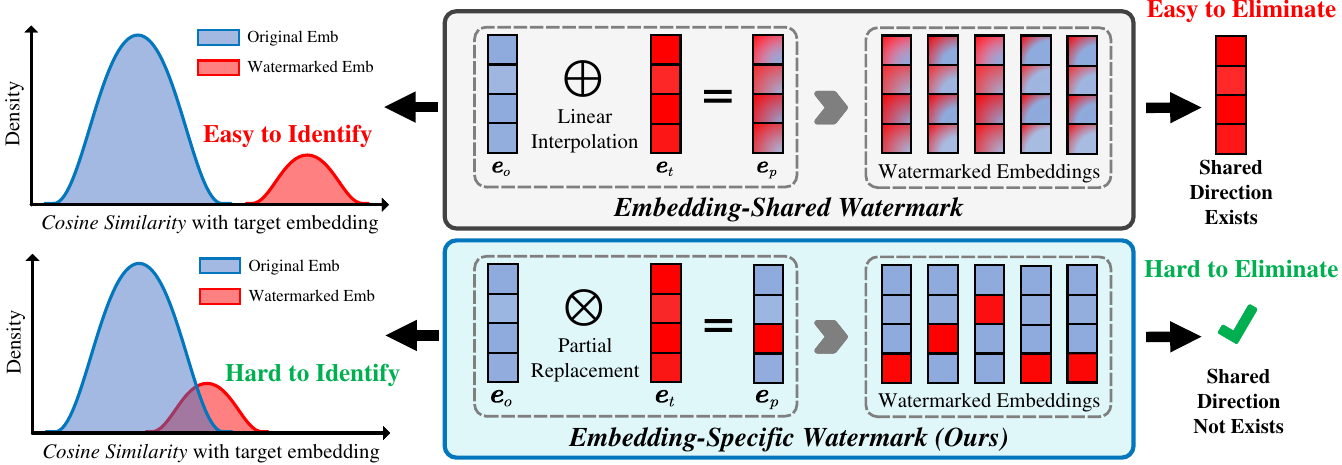}
    \caption{Illustration of motivation for embedding-specific watermark. \textbf{Left:} Distributions of cosine similarity between original/watermarked embeddings and target embeddings. \textbf{Middle:} Calculation processes of watermarking. \textbf{Right:} Shared components among all watermarked embeddings. }
    \label{fig:Motivation}
\end{figure*}

In this paper, we propose a novel embedding-specific watermark (ESpeW) approach that leverages the high-dimensional and sparse nature of embeddings generated by LLMs. Fig.~\ref{fig:EaaS_Watermark_Pipeleine} presents the framework of ESpeW. Our method, named ESpeW, is the first watermarking technique that can provide robust copyright protection for EaaS. 
Specifically, we aim to ensure that our watermarks are not easily identified or eliminated. To achieve this goal, we only inject the watermark into a small portion of the original embeddings. Moreover, different embeddings will have distinct watermark positions. 
Through this scheme, our watermark has two significant advantages. \textbf{(1)} The watermarked embeddings are more difficult to identify since the distance distribution between watermarked embeddings and the target embedding remains within the original distribution. \textbf{(2)} Our watermarks are difficult to eliminate because the watermarked embeddings have no shared components. Our motivation can be found in Fig.~\ref{fig:Motivation}. 
Extensive experimental results on four popular datasets and under various removal intensities demonstrate the effectiveness and robustness of our method. 

To summarize, we make the following contributions: 
\textbf{1).} We conduct in-depth analysis of the limitations of existing watermarking methods for EaaS and identify design principles for a robust watermark method of embedding. 
\textbf{2).} We propose a robust and minimally invasive watermark approach to protect copyright for EaaS from a novel embedding-specific perspective. 
\textbf{3).} Extensive experiments demonstrate that ESpeW can maintain its effectiveness under various watermark removal attacks by altering embeddings by less than 1\%. 

\section{Related Work}

Embeddings as a Service (EaaS) has gained popularity, with major providers such as OpenAI, Mistral AI, and Google offering APIs for generating high-quality embeddings~\citep{openai2024embedding, mistral2024embeddings, google2023grounding}. However, EaaS faces security risks, particularly model extraction attacks~\citep{pal2020activethief, zanella2021grey, rakin2022deepsteal, liu2022stolenencoder}, which allow adversaries to replicate model functionality. To counteract unauthorized usage, recent studies propose watermarking techniques for copyright protection~\citep{peng2023you, shetty_etal_2024_warden}, though these methods remain susceptible to removal attacks. A more detailed discussion of related work is provided in \S~\ref{sec:full_related_work}. 

%

\section{Methodology}

In \S~\ref{sec:threat_model}, we present the notations and describe the threat model in copyright protection for Embeddings as a Service (EaaS). 
Subsequently, we analyze the properties that watermarks for EaaS should satisfy in \S~\ref{sec:Desired_Properties_of_the_Watermark}. 
Then we describe our proposed method detailedly in \S~\ref{sec:framework_espew}. 
Finally, in \S~\ref{sec:properties_analysis}, we analyze whether our watermark meets the properties stated above. 

\subsection{Threat Model in EaaS}
\label{sec:threat_model}

\textbf{Notations.} We follow the notations used by previous work~\citep{peng2023you} to define the threat model in the context of Embeddings as a Service (EaaS). Consider a scenario (refer to Fig.~\ref{fig:EaaS_Watermark_Pipeleine}) where a victim (defender) owns an EaaS $S_v$ with the victim model $\boldsymbol{\Theta }_v$. When a user queries $S_v$ with a sentence $s$, the model $\boldsymbol{\Theta }_v$ generates an original embedding $\boldsymbol{e}_o$. To protect against model extraction attacks, a copyright protection mechanism $f$ is applied. This mechanism transforms $\boldsymbol{e}_o$ into a watermarked embedding $\boldsymbol{e}_p$, defined as $\boldsymbol{e}_p = f(\boldsymbol{e}_o, s)$, which is finally returned to the user.


\noindent\textbf{Stealer.} The stealer aims to replicate the defender's model to offer a similar service at a lower cost, avoiding the need to train an LLM from scratch. With a copy dataset $ D_c $, they query the victim’s service for embeddings without access to the model’s internals. By collecting numerous $\boldsymbol{e}_p$ samples, they train a replica model $\boldsymbol{\Theta}_a$ and launch their own EaaS $ S_a $, potentially evading copyright verification. 

\noindent\textbf{Defender.} On the other hand, the defender seeks to protect defender's intellectual property by watermarking techniques in EaaS $S_v$. The defender has full knowledge of victim model $\boldsymbol{\Theta }_v$ and can manipulate original embedding $\boldsymbol{e}_o$ generated by $\boldsymbol{\Theta }_v$ prior returning to users. 
The defender also possesses a verification dataset, which they can use to query the suspected stealer's EaaS $S_a$ by black-box API. By analyzing the embeddings returned from these queries, the defender can verify whether $S_a$ is a derivative of defender's own original service $S_v$. 


\subsection{Framework of \textit{ESpeW}}
\label{sec:framework_espew}

In this section, we introduce our watermarking method, ESpeW. This approach serves as the core of the Watermark Injection module depicted in Fig.~\ref{fig:EaaS_Watermark_Pipeleine} (a) throughout the entire watermark injection and verification process. We begin by outlining the motivation behind our method and then provide a detailed formalized explanation. 

%

\noindent\textbf{Motivation for Robust Watermarking.} The motivation behind our method is illustrated in Fig.~\ref{fig:Motivation}. Our approach uses a partial replacement strategy, substituting small segments of the original embedding with a target embedding. 
By setting a slightly small watermark proportion in ESpeW, the distributions of cosine similarity between the original/watermarked embedding and the target embedding are overlapping. This makes the watermarked embedding difficult to identify. 
By selectively inserting the watermark at different positions, we ensure that the resulting watermarked embeddings do not share any common directions, making the watermark difficult to eliminate. Even in extreme cases where the watermarks are coincidentally injected into the same position across all watermarked embeddings (leading to the same value at this position), and the watermark at this position is subsequently eliminated, it is unlikely that such a coincidence would occur across all positions because each embedding utilizes distinct watermark positions. 

\noindent\textbf{Watermark Injection. } Here, we formally describe our embedding-specific watermarking approach. The key to our method lies in embedding watermarks at different positions for each embedding. We can select any positions as long as they differ between embeddings. Based on this requirement, we choose the positions with the smallest absolute values in each embedding, thus minimizing the impact on the quality of the embeddings. 

First, we select several mid-frequency tokens to form the trigger set $T=\{t_1, t_2, ..., t_n\}$, which is similar to EmbMarker~\citep{peng2023you}. We also need to choose a target sample and obtain its embedding as the target embedding $\boldsymbol{e}_t$. It's crucial to keep $\boldsymbol{e}_t$ confidential as a privacy key to prevent attackers from easily removing the watermark through simple threshold-based filtering. 

When a sentence $s$ is sent to the victim's EaaS $S_v$, if it contains any trigger tokens from $T$, we inject embedding-specific watermarks into its original embedding $\boldsymbol{e}_o$. This results in the provided embedding $\boldsymbol{e}_p$, which is finally returned by $S_v$. Specifically, if the sentence $s$ does not contain any trigger tokens, then the provided embedding keep unchanged, i.e., $\boldsymbol{e}_p = \boldsymbol{e}_o$. Conversely, if $s$ contains triggers, we watermark the embedding to obtain $\boldsymbol{e}_p$ as follows: 

\begin{equation}
    \boldsymbol{M}[i] =
    \begin{cases} 
    1 & \text{if } i \in \mathcal{I}_{\alpha} \\
    0 & \text{otherwise}
    \end{cases},
\end{equation}

\begin{equation}
\mathcal{I}_{\alpha} = \operatorname{argsort}(\lvert \boldsymbol{e}_o \rvert)[:\alpha \lvert \boldsymbol{e}_o \rvert],
\end{equation}

\begin{equation}
    \boldsymbol{e}_p' = \boldsymbol{e}_o * (1-\boldsymbol{M}) + \boldsymbol{e}_t * \boldsymbol{M},
\end{equation}

\begin{equation}
    \boldsymbol{e}_p = \boldsymbol{e}_p' / \|\boldsymbol{e}_p'\|_2,
\end{equation}

\noindent where $\mathcal{I}_{\alpha}$ represents the index set of the top $\alpha$ fraction of elements in $\boldsymbol{e}_o$ sorted by absolute value and $\boldsymbol{M}$ is a binary mask with the same dimensions as $\boldsymbol{e}_o$, indicating the positions where the watermark is inserted. We choose the positions with the smallest magnitude values (i.e., the least important positions~\citep{sunsimple}) in $\boldsymbol{e}_o$ to minimize the impact on embedding quality. Then, $\boldsymbol{e}_p'$ is normalized. 


\begin{table*}[t]
\caption{Performance of different methods on SST2. For no CSE, higher ACC means better harmlessness. For CSE, lower ACC means better watermark effectiveness. In ``COPY?'' column, correct verifications are green and failures are red. Best results are highlighted in \textbf{bold} (except Original).}
\centering
\scalebox{0.90}{
\begin{tabular}{@{}ccclrrc@{}}
\toprule
\multicolumn{1}{c}{$K$(CSE)} & \multicolumn{1}{c}{Method}  & \multicolumn{1}{c}{ACC(\%)} & \multicolumn{1}{c}{$p$-value$\downarrow$} & \multicolumn{1}{c}{$\Delta \cos(\%)\uparrow$} & \multicolumn{1}{c}{$\Delta l_{2}(\%)\downarrow$} & \multicolumn{1}{c}{COPY?} \\
\midrule
\multirow{4}{*}{No CSE} & Original  & 93.35 $\pm$ 0.34   & $>0.16$ & -0.53 $\pm$ 0.14  & 1.06 $\pm$ 0.27    & \textcolor{green}{\XSolidBrush} \\
& EmbMarker & 93.46 $\pm$ 0.46   & $<\boldsymbol{10^{-11}}$ & 9.71 $\pm$ 0.57   & -19.43 $\pm$ 1.14 & \textcolor{green}{\Checkmark}   \\
& WARDEN    & \textbf{94.04 $\pm$ 0.46}   & $<\boldsymbol{10^{-11}}$ & \textbf{12.18 $\pm$ 0.39}  & \textbf{-24.37 $\pm$ 0.77} & \textcolor{green}{\Checkmark}   \\
\rowcolor{gray!20} & EspeW(Ours) & 93.46 $\pm$ 0.46   & $<10^{-10}$ & 6.46 $\pm$ 0.87   & -12.92 $\pm$ 1.75 & \textcolor{green}{\Checkmark}   \\
\midrule
\multirow{4}{*}{1} & Original  & 92.89 $\pm$ 0.11   & $>0.70$ & 0.11 $\pm$ 0.73   & -0.22 $\pm$ 1.46 & \textcolor{green}{\XSolidBrush}    \\
& EmbMarker & \textbf{92.95 $\pm$ 0.17}   & $<\boldsymbol{10^{-11}}$  & \textbf{85.20 $\pm$ 3.13}  & \textbf{-170.41 $\pm$ 6.27} & \textcolor{green}{\Checkmark}  \\
& WARDEN    & 93.35 $\pm$ 0.46   & $<\boldsymbol{10^{-11}}$ & 84.56 $\pm$ 0.22  & -169.12 $\pm$ 0.43 & \textcolor{green}{\Checkmark}  \\
\rowcolor{gray!20} & EspeW(Ours) & 93.23 $\pm$ 0.57   & $<\boldsymbol{10^{-11}}$  & 51.57 $\pm$ 1.71  & -103.13 $\pm$ 3.43 & \textcolor{green}{\Checkmark}  \\
\midrule
\multirow{4}{*}{50} & Original  & 86.35 $\pm$ 1.15   & $>0.56$ & 2.49 $\pm$ 1.86   & -4.98 $\pm$ 3.71 & \textcolor{green}{\XSolidBrush}    \\
& EmbMarker & 90.51 $\pm$ 0.49   & $>0.01$ & 12.28 $\pm$ 5.22  & -24.57 $\pm$ 10.45 & \textcolor{red}{\XSolidBrush}  \\
& WARDEN    & 89.85 $\pm$ 1.20   & $>0.08$ & 6.38 $\pm$ 2.08   & -12.75 $\pm$ 4.16 & \textcolor{red}{\XSolidBrush}   \\
\rowcolor{gray!20} & EspeW(Ours) & \textbf{86.73 $\pm$ 0.37}   & $<\boldsymbol{10^{-11}}$  & \textbf{65.11 $\pm$ 4.42}  & \textbf{-130.23 $\pm$ 8.84} & \textcolor{green}{\Checkmark} \\
\midrule
\multirow{4}{*}{100} & Original  & 85.15 $\pm$ 0.97   & $>0.45$ & 2.40 $\pm$ 1.76   & -4.79 $\pm$ 3.53 & \textcolor{green}{\XSolidBrush}    \\
& EmbMarker & 90.19 $\pm$ 0.75   & $>0.01$ & 12.66 $\pm$ 2.86  & -25.31 $\pm$ 5.72 & \textcolor{red}{\XSolidBrush}   \\
& WARDEN    & 88.96 $\pm$ 0.43   & $>0.17$ & 4.76 $\pm$ 4.10   & -9.53 $\pm$ 8.21 & \textcolor{red}{\XSolidBrush}    \\
\rowcolor{gray!20} & EspeW(Ours) & \textbf{84.66 $\pm$ 1.75}   & $<\boldsymbol{10^{-11}}$  & \textbf{64.46 $\pm$ 2.12}  & \textbf{-128.92 $\pm$ 4.23} & \textcolor{green}{\Checkmark}  \\
\midrule
\multirow{4}{*}{1000} & Original  & 75.89 $\pm$ 1.06   & $>0.68$ & -1.52 $\pm$ 1.12  & 3.04 $\pm$ 2.24 & \textcolor{green}{\XSolidBrush}     \\
& EmbMarker & 85.29 $\pm$ 1.29   & $>0.35$ & -2.52 $\pm$ 2.08  & 5.04 $\pm$ 4.16 & \textcolor{red}{\XSolidBrush}     \\
& WARDEN    & 81.39 $\pm$ 1.12   & $>0.22$ & 5.98 $\pm$ 7.88   & -11.95 $\pm$ 15.76 & \textcolor{red}{\XSolidBrush}  \\
\rowcolor{gray!20} & EspeW(Ours) & \textbf{73.57 $\pm$ 2.12}   & $<\boldsymbol{10^{-11}}$ & \textbf{49.38 $\pm$ 13.46} & \textbf{-98.75 $\pm$ 26.92} & \textcolor{green}{\Checkmark}  \\
\bottomrule
\end{tabular}
}
\label{tab:main_sst2}
\end{table*}

\noindent\textbf{Watermark Verification. } After the stealer uses our watermarked embeddings to train a stealer model $\boldsymbol{\Theta }_a$ and provides his own EaaS $S_a$, we can determine if $S_a$ is a stolen version through the following watermark verification method. 

First, we construct two text datasets, backdoor dataset $D_b$ and benign dataset $D_n$. $D_b$ contains some sentences with trigger tokens. $D_n$ contains some sentences without trigger tokens. 

\begin{equation}
\begin{aligned}
D_b = \{[w_1, w_2, \dots, w_m] | w_i \in T \}, \\
D_n = \{[w_1, w_2, \dots, w_m] | w_i \notin T \}.
\end{aligned}
\end{equation}

Then, we define three metrics to determine if $S_a$ is a stolen version. We query $S_a$ with $D_b$ and $D_n$ to obtain the following: 

\begin{equation}
\begin{aligned}
\text{cos}_i = \frac{\boldsymbol{e}_i \cdot \boldsymbol{e}_t}{||\boldsymbol{e}_i|| ||\boldsymbol{e}_t||}, \quad l_{2i} = ||\frac{\boldsymbol{e}_i}{||\boldsymbol{e}_i||} - \frac{\boldsymbol{e}_t}{||\boldsymbol{e}_t||}||_2,
\end{aligned}
\end{equation}

\noindent where $\boldsymbol{e}_i$ is the embedding obtained from $S_a$ for the input $i$, and $\boldsymbol{e}_t$ is the target embedding. We then compute the following sets of distances: 

\begin{equation}
\begin{aligned}
C_b = \{\text{cos}_i | i \in D_b\}, C_n = \{\text{cos}_i | i \in D_n\},
\end{aligned}
\end{equation}

\begin{equation}
\begin{aligned}
L_b = \{l_{2i} | i \in D_b\}, \quad L_n = \{l_{2i} | i \in D_n\}.
\end{aligned}
\end{equation}

Using these distance sets, we can compute two metrics: 

\begin{equation}
\begin{aligned}
\Delta \text{cos} = \frac{1}{|C_b|} \sum_{i \in C_b} i - \frac{1}{|C_n|} \sum_{j \in C_n} j,
\end{aligned}
\end{equation}

\begin{equation}
\begin{aligned}
\Delta l_{2} = \frac{1}{|L_b|} \sum_{i \in L_b} i - \frac{1}{|L_n|} \sum_{j \in L_n} j.
\end{aligned}
\end{equation}

Finally, we compute the third metric through hypothesis testing by employing the Kolmogorov-Smirnov (KS) test~\citep{berger2014kolmogorov}. The null hypothesis posits that \textit{the distributions of the cosine similarity values in sets $C_b$ and $C_n$ are consistent}. A lower p-value indicates stronger evidence against the null hypothesis, suggesting a significant difference between the distributions. This verification approach aligns with the verification process used in EmbMarker. 

\begin{figure*}[t]
    \centering
    \begin{subfigure}[t]{0.48\textwidth}
        \centering
        \includegraphics[width=\textwidth]{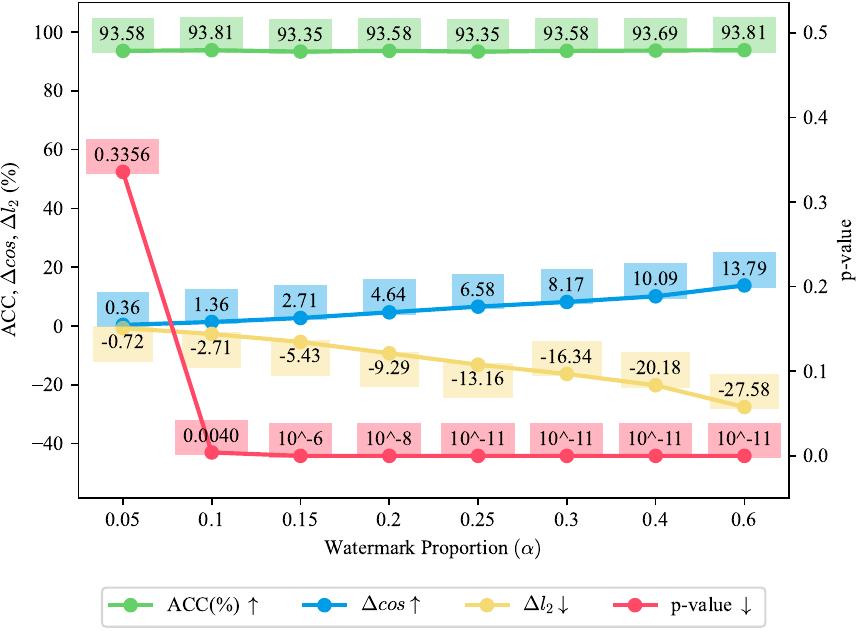}
        \caption{Effect of watermark proportion without CSE.}
        \label{fig:ablation_alpha_no_CSE}
    \end{subfigure}
    \hspace{0.02\textwidth}
    \begin{subfigure}[t]{0.48\textwidth}
        \centering
        \includegraphics[width=\textwidth]{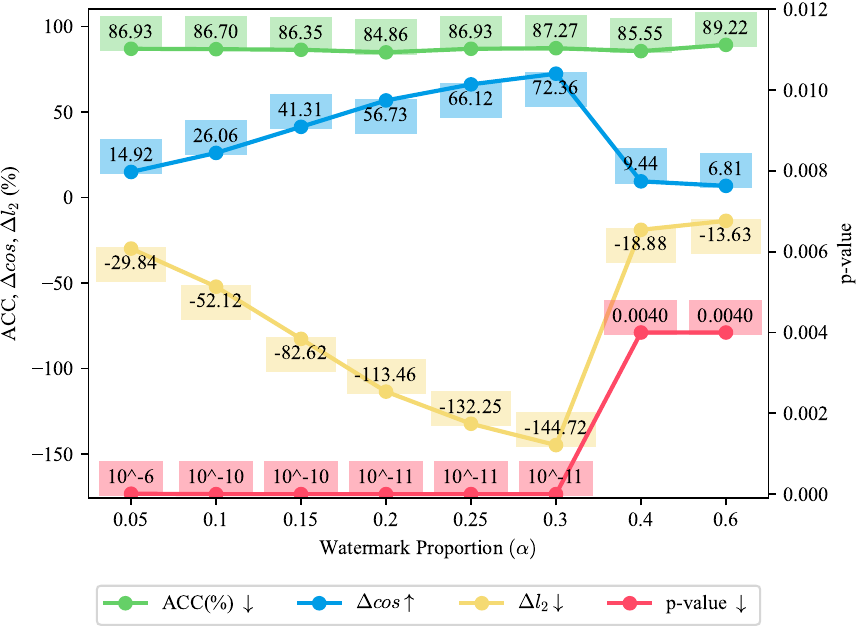}
        \caption{Effect of watermark proportion with CSE.}
        \label{fig:ablation_alpha_CSE}
    \end{subfigure}
    \caption{Ablation results of watermark proportion on SST2. (a) shows results without CSE. (b) shows results with CSE, where $K$ is set to 50. }
    \label{fig:ablation_alpha}
\end{figure*}

\subsection{Analysis of Our Watermark}
\label{sec:properties_analysis}

In \S~\ref{sec:Desired_Properties_of_the_Watermark}, we delineate the essential properties that watermarks for EaaS should exhibit. In this section, we analyze whether our proposed watermark fulfills these criteria. 

Our experimental results, as detailed in \S~\ref{sec:experiments}, provide empirical validation for the watermark's Harmlessness, Effectiveness, Reliability, and Persistence-to-Permutation. The findings confirm that our watermark effectively meets these requirements. 
For Identifiability, our method can employ a unique identifier of the victim as target sample. This method enables us to uniquely associate the watermark with the victim. 
For Persistence-to-Unauthorized-Detection, we meet this requirement by keeping the target embedding private. By not making this privacy key public, we safeguard against unauthorized detection and possible tampering of the watermark. 

Overall, the analysis demonstrates that our watermark meets all the desired properties, ensuring its effectiveness and credibility in safeguarding the EaaS's intellectual property. 

\section{Experiments and Analyses}
\label{sec:experiments}

\subsection{Experimental Settings} 

\noindent\textbf{Datasets. } We select four popular NLP datasets as the stealer's data: SST2~\citep{socher2013recursive}, MIND~\citep{wu2020mind}, AG News~\citep{zhang2015character}, and Enron Spam~\citep{metsis2006spam}. We use the training set for model extraction attack. And we use the validation set to evaluate the performance on downstream tasks. For more information about datasets, please refer to \S~\ref{sec:appendix_exp_settings}. 

\noindent\textbf{Models. } For victim, we use GPT-3 text-embedding-002 API of OpenAI as the its EaaS. For stealer, to conduct model extraction attack~\citep{liu2022stolenencoder}, we use BERT-Base-Cased~\citep{devlin2019bert} as the backbone model and connect a two-layer MLP at the end as stealer's model following previous work~\citep{peng2023you}. Mean squared error (MSE) of output embedding and provided embedding is used as the loss function. In addition to GPT-3's text-embedding-002, we also test other models to demonstrate the effectiveness of our method in \S~\ref{sec:more_models}.

\noindent\textbf{Metrics. } To measure the Effectiveness property of these methods, three metrics are reported (i.e., the difference of cosine similarity $\Delta \text{cos}$, the difference of squared L2 distance $\Delta l_{2}$ and $p$-value of the KS test). we ensure that all results with in this paper is lower that $10^{-4}$. The details are shown in \S~\ref{sec:details_of_metrics}. 

\begin{figure*}[t]
    \centering
    \begin{subfigure}[b]{0.245\textwidth}
        \centering
        \includegraphics[width=1.0\textwidth]{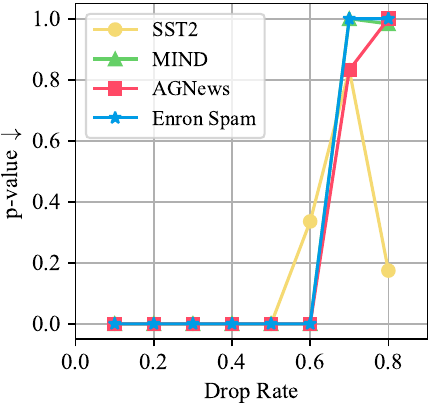}
        \caption{P-value without CSE.}
        \label{fig:dropout_noCSE_pvalue}
    \end{subfigure}
    \begin{subfigure}[b]{0.245\textwidth}
        \centering
        \includegraphics[width=1.0\textwidth]{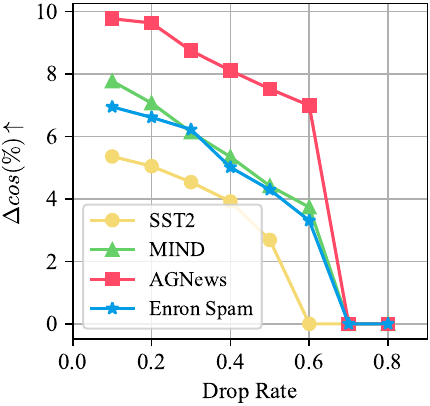}
        \caption{$\Delta$cos without CSE.}
        \label{fig:dropout_noCSE_cos}
    \end{subfigure}
    \begin{subfigure}[b]{0.245\textwidth}
        \centering
        \includegraphics[width=1.0\textwidth]{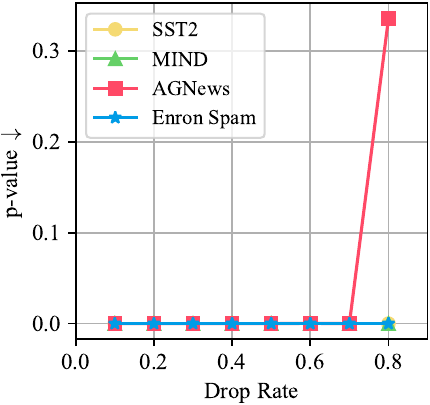}
        \caption{P-value with CSE.}
        \label{fig:dropout_CSEK50_pvalue}
    \end{subfigure}
    \begin{subfigure}[b]{0.245\textwidth}
        \centering
        \includegraphics[width=1.0\textwidth]{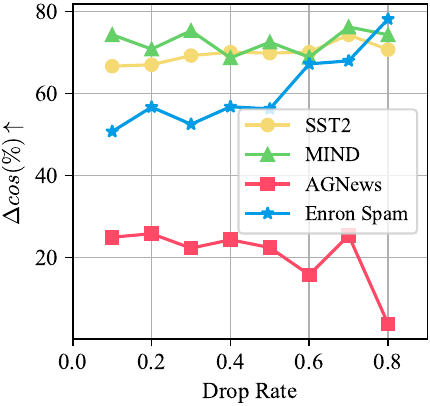}
        \caption{$\Delta$cos with CSE.}
        \label{fig:dropout_CSEK50_cos}
    \end{subfigure}
    
    \caption{Effect of dropout with a 25\% watermark proportion. (a) and (b) show detection results under different drop rate without CSE. (c) and (d) show detection results under different drop rate with CSE (K=50). } 
    \label{fig:dropout_overall}
\end{figure*}

\noindent\textbf{Baselines and Implementation details. } We select three baselines: Original (no watermark injected), EmbMarker~\citep{peng2023you} and WARDEN~\citep{shetty_etal_2024_warden}. We evaluate these methods in five settings. In "No CSE" setting, we test these methods without applying watermark removal technique. Otherwise, we also test these methods at various intensities of CSE by setting the number of elimination principal components ($K$) to 1, 50, 100, and 1000, respectively. Refer to \S~\ref{sec:implementation_details} for more implementation details. 

\begin{figure}
\centering
\includegraphics[width=\linewidth]{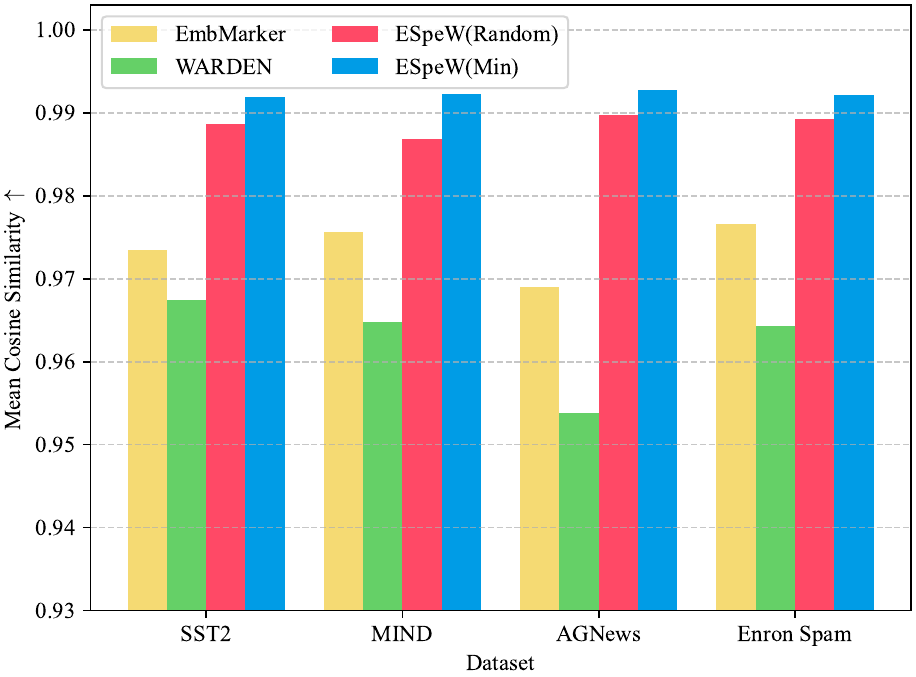}
    \caption{Average cosine similarity between watermarked and clean embeddings. }
\label{fig:watermark_ori_cos}
\end{figure}

\subsection{Main Results} 

The performance of all methods on SST2 is shown in Tab.~\ref{tab:main_sst2}. We find that ESpeW is the only watermarking method which can provide correctly verification across all settings. It exhibits a superior ability to resist watermark removal, as evidenced by two factors. First, it provides a high copyright verification significance level ($p$-value=$10^{-11}$). Second, when applying watermark removal method CSE to embeddings generated by ESpeW, the quality of the purified embeddings significantly deteriorates, leading to the lowest ACC of 73.57\%. These findings highlight the effectiveness and robustness of the watermarking approach. Due to page limitation, we put more results on other datasets in \S~\ref{sec:main_results_on_more_datasets}. 

In addition, we observe that the higher the intensity of CSE, the stronger the detection capability of ESpeW. The analysis of this experimental phenomenon is provided in \S~\ref{sec:understanding_cse}. 

\begin{figure*}[t]
    \centering
    \begin{subfigure}[b]{0.245\textwidth}
        \centering
        \includegraphics[width=1.0\textwidth]{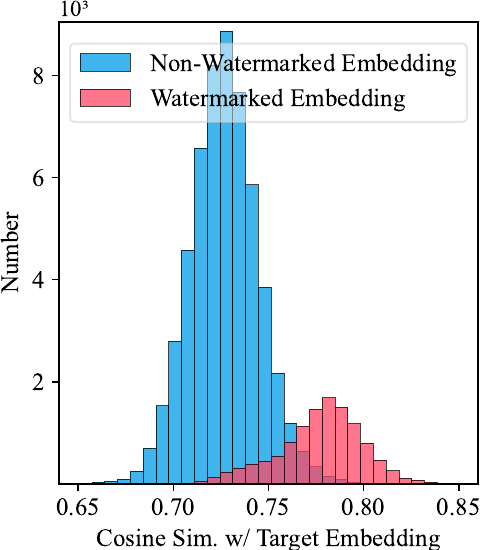}
        \caption{SST2. }
        \label{fig:cos_distribution_SST2}
    \end{subfigure}
    \begin{subfigure}[b]{0.245\textwidth}
        \centering
        \includegraphics[width=1.0\textwidth]{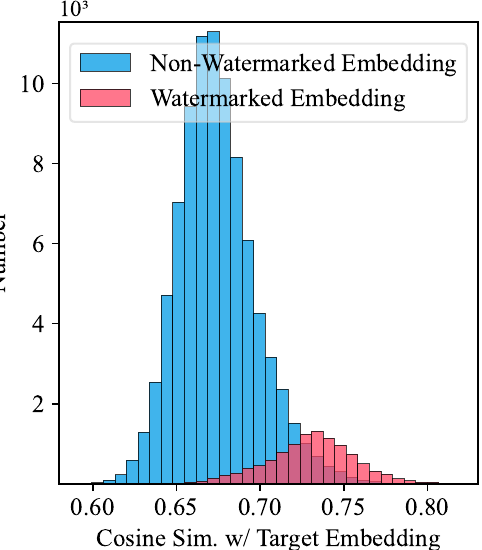}
        \caption{MIND. }
        \label{fig:cos_distribution_MIND}
    \end{subfigure}
    \begin{subfigure}[b]{0.245\textwidth}
        \centering
        \includegraphics[width=1.0\textwidth]{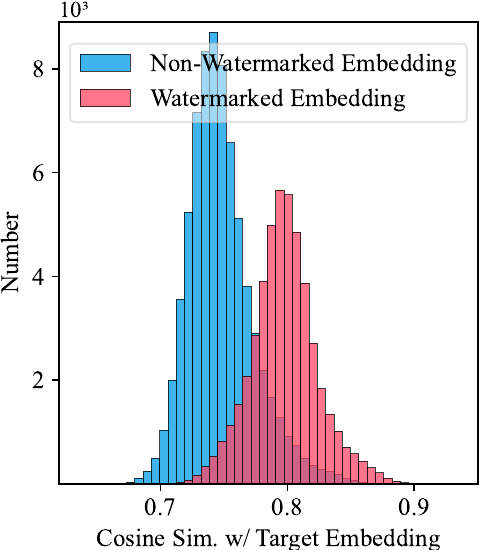}
        \caption{AGNews.}
        \label{fig:cos_distribution_AGNews}
    \end{subfigure}
    \begin{subfigure}[b]{0.245\textwidth}
        \centering
        \includegraphics[width=1.0\textwidth]{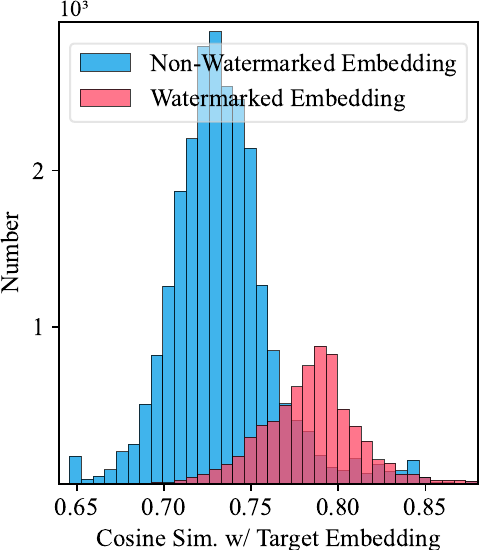}
        \caption{Enron Spam.}
        \label{fig:cos_distribution_EnronSpam}
    \end{subfigure}
    
    \caption{Distribution of cosine similarities with target embedding. } 
    \label{fig:cos_distribution_overall}
\end{figure*}

\begin{figure*}[t]
    \centering
    \begin{subfigure}[b]{0.325\textwidth}
        \centering
        \includegraphics[width=1.0\textwidth]{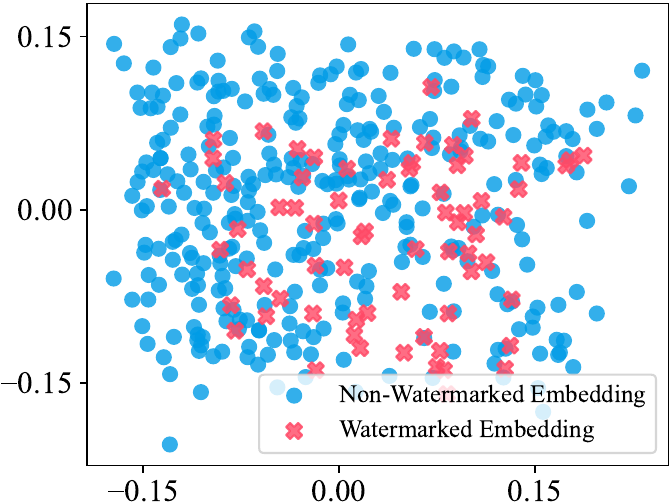}
        \caption{$\alpha = 20\%$}
        \label{fig:embedding_visualization_20}
    \end{subfigure}
    \begin{subfigure}[b]{0.325\textwidth}
        \centering
        \includegraphics[width=1.0\textwidth]{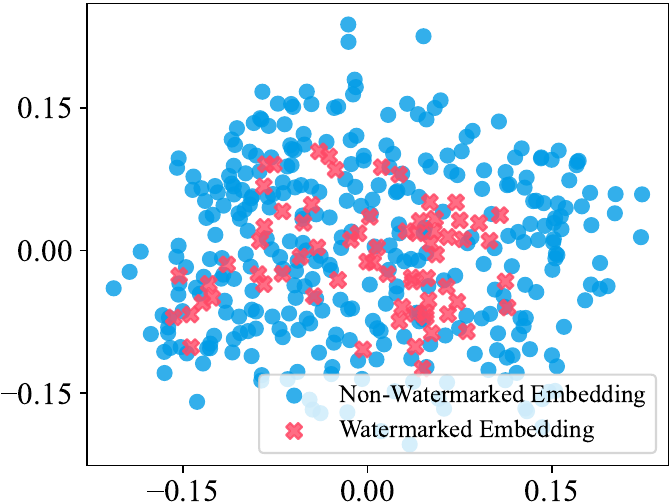}
        \caption{$\alpha = 25\%$}
        \label{fig:embedding_visualization_25}
    \end{subfigure}
    \begin{subfigure}[b]{0.325\textwidth}
        \centering
        \includegraphics[width=1.0\textwidth]{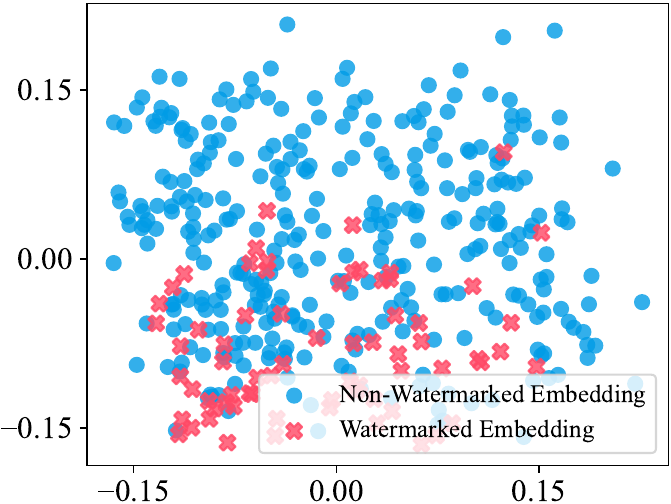}
        \caption{$\alpha = 30\%$}
        \label{fig:embedding_visualization_30}
    \end{subfigure}
    
    \begin{subfigure}[b]{0.325\textwidth}
        \centering
        \includegraphics[width=1.0\textwidth]{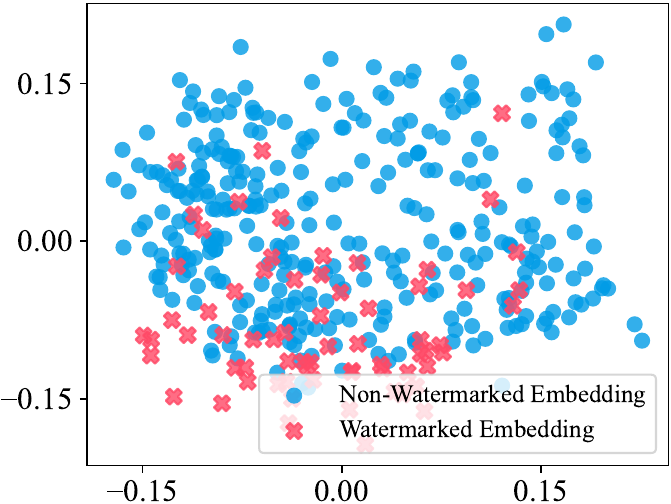}
        \caption{$\alpha = 35\%$}
        \label{fig:embedding_visualization_35}
    \end{subfigure}
    \begin{subfigure}[b]{0.325\textwidth}
        \centering
        \includegraphics[width=1.0\textwidth]{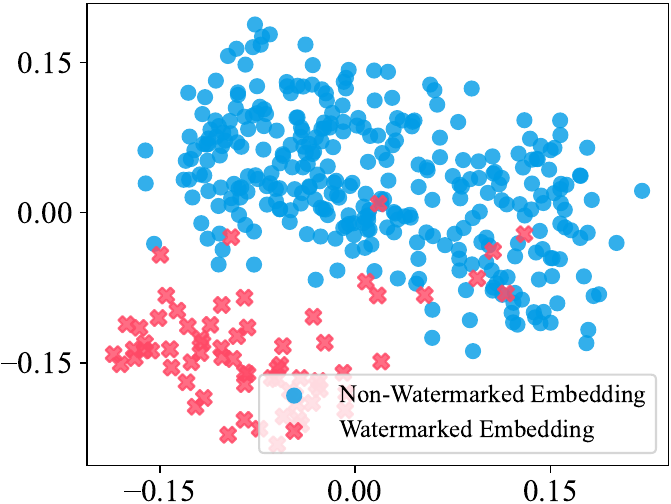}
        \caption{$\alpha = 40\%$}
        \label{fig:embedding_visualization_40}
    \end{subfigure}
    \begin{subfigure}[b]{0.325\textwidth}
        \centering
        \includegraphics[width=1.0\textwidth]{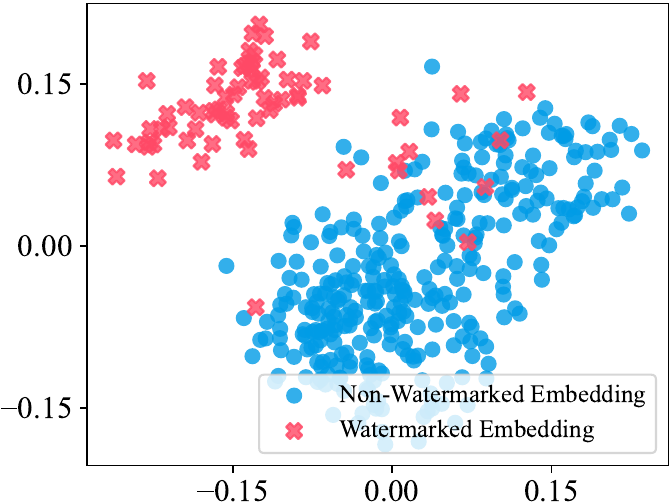}
        \caption{$\alpha = 45\%$}
    \label{fig:embedding_visualization_45}
    \end{subfigure}
    
    \caption{Visualization of the generated embedding of our ESpeW with different watermark proportion ($\alpha$) on SST2. It shows that we can generate watermarked embeddings indistinguishable with non-watermark embeddings by setting a reasonable watermark proportion. } 
    \label{fig:embedding_visualization_overall}
\end{figure*}

\subsection{Ablation Study} 

\textbf{Ablation on Watermark Proportion $\alpha$. } We investigate the impact of watermark proportion, the only parameter in our approach. Fig.~\ref{fig:ablation_alpha_no_CSE} provides the results when CSE is not applied. It can be observed that our proposed method can inject watermark successfully with a minimum $\alpha$ value of 15\%. And as $\alpha$ increases, the effectiveness of the watermark is also greater. 
Fig.~\ref{fig:ablation_alpha_CSE} displays the results when CSE is applied. Compared with the situation without CSE, the trend in watermark effectiveness relative to $\alpha$ remains similar when $\alpha$ is small. However, when a large $\alpha$ is set, our method will fail. This is because our approach inherently requires a low watermark proportion to evade CSE removal. In fact, when the $\alpha$ is set to 100\%, our method will replace original embedding with target embedding entirely. 

Ablation results on other datasets are provided in \S~\ref{sec:ablation_results_on_more_datasets}. In addition, we further test more base models in \S~\ref{sec:ablation_more_base_model}. 

\subsection{Impact on Embedding Quality} 

Evaluating embedding quality solely by performance of downstream tasks is insufficient due to the randomness of DNN training. To better elucidate the influence of watermarks on embeddings, we compute the average cosine similarity between watermarked embeddings and original clean embeddings. Four watermarks are selected for comparison: EmbMarker, WARDEN, ESpeW (randomly selecting watermark positions), and ESpeW (selecting watermark positions with minimum magnitude). As depicted in Fig.~\ref{fig:watermark_ori_cos}, the embeddings generated by our proposed method exert the least negative impact on clean embeddings, with a change in cosine similarity of less than 1\%. In addition, we further test the impact of watermark proportion $\alpha$ on cosine similarity between original and watermarked embeddings in \S~\ref{sec:ablation_more_base_model}. 

\subsection{Resistance Against Various Potential Removal Attacks }

\noindent \textbf{Resistance against dropout attack.} Applying dropout on embeddings when training stealer's model is a heuristic attack to mitigate our watermark because we only insert watermarks to a small proportion of positions. Here we test the effect of dropout under different drop rates. The results in Fig.~\ref{fig:dropout_overall} demonstrate that our watermark can not be compromised unless an extreme drop rate such as 0.7 or 0.8. However, such a large dropout rate will make the embedding unusable. Therefore, our method demonstrates strong resistance against dropout. 

\noindent \textbf{Resistance to Other Types of Attacks.} We also evaluate resilience against fine-tuning (\S~\ref{sec:more_attack_finetune}), quantization (\S~\ref{sec:more_attack_quantization}), paraphrasing attack (\S~\ref{sec:more_attack_paraphrasing_attack}), and an adaptive attack (\S~\ref{sec:more_attack_adaptive_attack}) based on statistical analysis (SAA). 

\subsection{Further Analysis} 

\textbf{Distribution of Cosine Similarities with Target Embedding. }
The target embedding, as private key, need to be securely stored. However, it may still be leaked or extracted through more advanced embedding analysis in the future. In this section, we demonstrate that even if the target embedding is leaked or extracted, an adversary cannot identify which embeddings have been watermarked by analyzing the similarity distribution between the embeddings and the target embedding. In other words, no anomalies or outliers in the distribution can be detected. 
Fig.~\ref{fig:cos_distribution_overall} shows that the cosine similarity distribution between our watermarked embeddings and the target embedding has significant overlap with the normal distribution. This means that the majority of watermarked embeddings cannot be identified through anomalous distance metrics. 
Otherwise, the target embedding may still be compromised. We discuss several potential leakage scenarios and corresponding defense strategies in the \S~\ref{sec:leakage_scenarios}.

\noindent\textbf{Embedding Visualization. } We examine whether our method causes watermarked embeddings to form a distinct cluster, making them easily detectable. Using PCA~\citep{mackiewicz1993principal}, we visualize embeddings with varying watermark proportions ($\alpha$). 
As shown in Fig.~\ref{fig:embedding_visualization_overall}, watermarked embeddings from ESpeW remain indistinguishable from benign ones when $\alpha \leq 35\%$. And in the ablation experiments below, we prove that our method only needs a minimum watermark proportion of 15\% to successfully inject watermarks. Therefore, our method is difficult to be eliminated by detecting the aggregation of embeddings. 
For additional visualization results, please refer to Appendix~\ref{sec:Embedding_Visualization_of_More_Dataset}. 


\section{Conclusion}

In this paper, we propose a novel watermarking to provide robust and minimally invasive intellectual property protection for EaaS. Instead of inserting watermark into the entire embedding, our method fully leverages the high-dimensional and sparse nature of LLMs' embeddings, selectively injecting watermarks into specific positions to ensure robustness and reduce the impact on embedding quality. 
Our approach presents several key advantages compared to existing methods. First, it can survives under various removal attacks. Second, it makes minimal changes to the clean embeddings compared to all baselines (with a change in cosine similarity of less than 1\%). Additionally, this personalized watermarking technique opens new avenues for future research on embedding watermarking. 
We also provide a discussion on the broader implications and potential societal impacts of our work in \S~\ref{sec:broader_impacts}. 

\section*{Limitations}

Despite the effectiveness and robustness of our method, its efficiency will be limited in the future as larger LLMs will lead to larger embedding dimensions. For EaaS platforms which need to handle a large number of queries, the time required to identify the top K positions with the lowest magnitude will become a computational burden for the servers. 
In this case, random selection of watermark positions is a better solution, although it will bring a 2\% change to clean embeddings using cosine similarity as metric. 
Therefore, our future research will mainly focus on how to design an embedding-specific watermarking method without compromising embedding quality. Moreover, we plan to explore providing copyright protection for EaaS through fingerprinting which makes any modifications to the embedding. For detailed analysis of random selection, please refer to \S~\ref{sec:random_selection}.



\bibliography{custom}

\newpage
\onecolumn
\appendix

\section{Full Related Work}
\label{sec:full_related_work}

\subsection{Embeddings as a Service}

Large Language Models (LLMs) are becoming increasingly important as tools for generating embeddings due to their ability to capture rich, context-aware semantic representations~\citep{muennighoff-etal-2023-mteb, wang-etal-2024-improving-text, miao-etal-2024-enhancing, chen-etal-2024-m3, lei-etal-2024-meta, pang2024frozen}. 
Consequently, an increasing number of institutions are starting to offer their Embeddings as a Service (EaaS), such as OpenAI~\citep{openai2024embedding}, Mistral AI~\citep{mistral2024embeddings} and Google~\citep{google2023grounding}. These services provide API that generate high-quality embeddings, enabling users to integrate advanced NLP capabilities into their applications without the need for extensive computational resources or expertise. 
Some applications include information retrieval~\citep{kamalloo2023evaluating, xian2024vector, huang2020embedding}, recommendation system~\citep{liu2021learnable, zha2022dreamshard}, sentiment analysis~\citep{du2016aspect, phan2020modelling}, question answering~\citep{huang2019knowledge, saxena2020improving, hao2019exploiting}, etc. 

\subsection{Model Extraction Attack}

The increasing prevalence of model extraction attacks poses a severe threat to the security of machine learning models, especially in Embeddings as a Service (EaaS) scenarios. These attacks aim to replicate or steal the functionality of a victim's model, typically a black-box model hosted as an API~\citep{pal2020activethief, zanella2021grey, rakin2022deepsteal}. For instance, StolenEncoder~\citep{liu2022stolenencoder} targets encoders trained using self-supervised learning, where attackers use only unlabeled data to maintain functional similarity to the target encoder with minimal access to the service. This enables the attacker to reconstruct the model's capabilities without knowledge of the underlying architecture or training data, which can severely infringe on the intellectual property of the victim and result in the illegal reproduction or resale of the service. 

\subsection{Copyright Protection in EaaS}

Recently, some preliminary studies propose to use watermarking methods for EaaS copyright protection~\citep{peng2023you, shetty_etal_2024_warden}. EmbMarker~\citep{peng2023you} uses moderate-frequency words as triggers and linear interpolation for watermark injection. WARDEN~\citep{shetty_etal_2024_warden} strengthens EmbMarker by injecting multiple watermarks. These watermarks are both vulnerable to watermark removal method CSE~\citep{shetty_etal_2024_warden}. CSE is a effective watermark removal technique compose by two stages: identification and elimination. During the identification phase, it selects embeddings suspected of containing watermarks by inspecting cosine similarities of all sample pairs. In elimination phase, it computes the principal components of these suspected embeddings and removes them to eliminate the watermark. Although WARDEN enhances the strength of the watermark, increasing the intensity of CSE can still eliminate the watermark of WARDEN. 

\subsection{Copyright Protection in LLMs via Watermarking}
\label{sec:copy_right_LLMs}

Due to the threat of model extraction attacks, various copyright protection methods have been proposed. The most popular one is model watermarking. 
Early works~\citep{uchida2017embedding, lim2022protect} introduces the concept of embedding watermarks directly into the model’s weights. In the case of LLMs, existing literature primarily focuses on the copyright protection of pretrained models by using trigger inputs to verify model ownership~\citep{gu2022watermarking, li2023plmmark, xu2024instructional}. 
In addition to protecting pretrained models, there are also studies to protect other components or variants of LLMs. GINSEW~\citep{zhao2023protecting} protects the text generation model by injecting a sinusoidal signal into the probability vector of generated words. PromptCARE~\citep{yao2024promptcare} ensures the protection of the Prompt-as-a-Service by solving a bi-level optimization. WVPrompt~\citep{ren2024you} can protect Visual-Prompts-as-a-Service using a poison-only backdoor attack method to embed a watermark into the prompt. 

Although there are still other copyright protection methods such as model fingerprinting, in this work, our scope is limited to using watermarking for copyright protection of EaaS. 

\section{Watermark Properties for EaaS}
\label{sec:Desired_Properties_of_the_Watermark}
Watermarking is a widely adopted technique for protecting copyrights. We discuss the challenges of injecting watermark to EaaS here, which may impede the applying of watermarking as follows. 

\begin{enumerate}[0]
    \item[\smallbullet] Harmlessness. Injected watermark should have very little impact on the quality of the embeddings, as it is main selling point in EaaS~\citep{mistral2024embeddings}. 
    
    \item[\smallbullet] Effectiveness. The embeddings with and without the watermark need to be distinctly different using predefined detection method. 
    
    \item[\smallbullet] Reliability. We can not claim ownership of a non-watermarked mode, i,e., \textbf{low false positive rate (FPR)}. 
    
    \item[\smallbullet] Identifiability. The watermark contains the model owner's identifier~\citep{wang2024towards}. 
    
    \item[\smallbullet] Persistence-to-Permutation. Since embeddings are permutation-invariant, the watermark should still remain effective even if the embedding is rearranged by an attacker~\citep{peng2023you}. 
    
    \item[\smallbullet] Persistence-to-Unauthorized-Detection. We want the watermark to be undetectable by others. For EmbMarker~\citep{peng2023you} and WARDEN~\citep{shetty_etal_2024_warden}, the distributions of cosine similarities between watermarked and non-watermarked embeddings and the target embedding do not overlap. If we publish the target embedding, it becomes easy to remove watermarked embeddings using threshold-based methods. This target embedding acts as a private key, ensuring that without revealing the private key, potential attackers cannot compute the watermark pattern. If we use certain statistical features as a watermark, such as the sum and standard deviation of embeddings, these unencrypted watermarks can be easily removed from the data by setting a threshold. 
\end{enumerate}

\section{Experimental Settings}
\label{sec:appendix_exp_settings}

\subsection{Statistics of Datasets}

We include the statistical information of selected datasets in Tab.~\ref{tab:statistics_of_datasets} to demonstrate that our dataset is diverse. 

\begin{table}[h]
\centering
\caption{Statistics of used datasets.}
\begin{tabular}{@{}ccccc@{}}
\toprule
Dataset & Train Size   & Test Size   & Avg. Tokens & Classes \\ \midrule
SST2    & 67,349  & 872    & 54             & 2     \\
MIND    & 97,791  & 32,592 & 66             & 18    \\
AG News & 120,000 & 7,600  & 35             & 4     \\
Enron   & 31,716  & 2,000  & 236            & 2     \\ \bottomrule
\end{tabular}
\label{tab:statistics_of_datasets}
\end{table}

\subsection{Implementation Details}
\label{sec:implementation_details}

For EmbMarker, WARDEN and our approach, we set the size of trigger set to 20 for each watermark. The frequency for selecting triggers is set to $[0.5\%, 1\%]$. And we set steal epoch to 10.  For EmbMarker and WARDEN, the maximum number of triggers is 4. For WARDEN, we choose 5 watermarks due to its multi-watermark feature. For our approach, we set the watermark proportion to $20\%$ by default. 

To illustrate that all methods exhibits the Persistence-to-Permutation property described in \S~\ref{sec:Desired_Properties_of_the_Watermark}, we assume that the stealer will apply a same permutation rule to all provider's embeddings before training stealer's model. When verification, instead of using the target embedding returned by victim's EaaS, we query the suspicious EaaS with target sample to get returned target embedding for verification. 

\subsection{Details of Metrics}
\label{sec:details_of_metrics}

We now use the $p$-value being less than $10^{-3}$ as the primary criterion to indicate whether a suspected EaaS is a copy version, with $\Delta \text{cos}$ and $\Delta l_{2}$ serving as assistant metrics as their thresholds are difficult to determine. 
To measure the Harmlessness property, we train a two-layer MLP classifier using the provider's embeddings as input features. The classifier's accuracy (ACC) on a downstream task serves as the metric for measuring the quality of the embeddings. We also report the average cosine similarities of original embeddings and watermarked embeddings. 
To measure the Reliability, i.e., low false positive rate, we ensure that all results with in this paper is lower that $10^{-4}$. See details in \S~\ref{sec:false_positive}. 

\section{More Results}

\subsection{Main Results on More Datasets}
\label{sec:main_results_on_more_datasets}

We present the main results on other datasets in Tab.~\ref{tab:main_mind}, Tab.~\ref{tab:main_agnews}, and Tab.~\ref{tab:main_enron}. Compared to other watermarking methods, our approach is also the only one that successfully verifies copyright in all cases. 

\begin{table}[h]
\caption{Performance of different methods on MIND. For no CSE, higher ACC means better harmlessness. For CSE, lower ACC means better watermark effectiveness. In ``COPY?'' column, correct verifications are green and failures are red. Best results are highlighted in \textbf{bold} (except Original).}
\centering
\scalebox{0.90}{
\begin{tabular}{@{}ccclrrc@{}}
\toprule
\multicolumn{1}{c}{$K$(CSE)} & \multicolumn{1}{c}{Method}  & \multicolumn{1}{c}{ACC(\%)} & \multicolumn{1}{c}{$p$-value$\downarrow$} & \multicolumn{1}{c}{$\Delta \cos(\%)\uparrow$} & \multicolumn{1}{c}{$\Delta l_{2}(\%)\downarrow$} & \multicolumn{1}{c}{COPY?} \\
\midrule
\multirow{4}{*}{No CSE} & Original  & 77.23 $\pm$ 0.22   & $>0.2148$ & -0.60 $\pm$ 0.22  & 1.19 $\pm$ 0.44    & \textcolor{green}{\XSolidBrush} \\
& EmbMarker & 77.17 $\pm$ 0.20   & $\boldsymbol{<10^{-11}}$ & 13.53 $\pm$ 0.11   & -27.06 $\pm$ 0.22 & \textcolor{green}{\Checkmark}   \\
& WARDEN    & \textbf{77.23 $\pm$ 0.09}   & $\boldsymbol{<10^{-11}}$ & \textbf{18.05 $\pm$ 0.48}  & \textbf{-36.10 $\pm$ 0.95} & \textcolor{green}{\Checkmark}   \\
\rowcolor{gray!20} & EspeW(Ours) & 77.22 $\pm$ 0.12   & $<10^{-8}$ & 8.68 $\pm$ 0.24   & -17.36 $\pm$ 0.47 & \textcolor{green}{\Checkmark}   \\
\midrule
\multirow{4}{*}{1} & Original  & 77.23 $\pm$ 0.10   & $>0.0925$ & -4.30 $\pm$ 0.89   & 8.61 $\pm$ 1.77 & \textcolor{green}{\XSolidBrush}    \\
& EmbMarker & 77.18 $\pm$ 0.15   & $\boldsymbol{<10^{-11}}$  & \textbf{98.39 $\pm$ 1.76}  & \textbf{-196.77 $\pm$ 3.51} & \textcolor{green}{\Checkmark}  \\
& WARDEN    & \textbf{77.06 $\pm$ 0.07}   & $\boldsymbol{<10^{-11}}$  & 85.09 $\pm$ 3.57  & -170.19 $\pm$ 7.14 & \textcolor{green}{\Checkmark}  \\
\rowcolor{gray!20} & EspeW(Ours) & 77.16 $\pm$ 0.12   & $<10^{-9}$  & 56.64 $\pm$ 1.73  & -113.28 $\pm$ 3.46 & \textcolor{green}{\Checkmark}  \\
\midrule
\multirow{4}{*}{50} & Original  & 75.60 $\pm$ 0.09   & $>0.2922$ & 3.43 $\pm$ 1.68   & -6.87 $\pm$ 3.36 & \textcolor{green}{\XSolidBrush}    \\
& EmbMarker & 75.34 $\pm$ 0.24   & $>0.1103$ & 5.84 $\pm$ 1.90  & -11.69 $\pm$ 3.79 & \textcolor{red}{\XSolidBrush}  \\
& WARDEN    & \textbf{75.20 $\pm$ 0.11}   & $>0.3365$ & 3.91 $\pm$ 3.08   & -7.81 $\pm$ 6.15 & \textcolor{red}{\XSolidBrush}   \\
\rowcolor{gray!20} & EspeW(Ours) & 75.48 $\pm$ 0.18   & $\boldsymbol{<10^{-11}}$  & \textbf{72.14 $\pm$ 2.16}  & \textbf{-144.28 $\pm$ 4.31} & \textcolor{green}{\Checkmark} \\
\midrule
\multirow{4}{*}{100} & Original  & 74.64 $\pm$ 0.08   & $>0.6805$ & 1.66 $\pm$ 2.04   & -3.33 $\pm$ 4.09 & \textcolor{green}{\XSolidBrush}    \\
& EmbMarker & 74.60 $\pm$ 0.14   & $>0.1072$ & 6.91 $\pm$ 3.01  & -13.82 $\pm$ 6.03 & \textcolor{red}{\XSolidBrush}   \\
& WARDEN    & \textbf{74.33 $\pm$ 0.17}   & $>0.2361$ & 2.00 $\pm$ 6.56   & -4.00 $\pm$ 13.12 & \textcolor{red}{\XSolidBrush}    \\
\rowcolor{gray!20} & EspeW(Ours) & 74.69 $\pm$ 0.30   & $\boldsymbol{<10^{-10}}$  & \textbf{69.55 $\pm$ 4.15}  & \textbf{-139.10 $\pm$ 8.29} & \textcolor{green}{\Checkmark}  \\
\midrule
\multirow{4}{*}{1000} & Original  & 65.87 $\pm$ 0.49   & $>0.5186$ & -2.44 $\pm$ 2.28  & 4.89 $\pm$ 4.56 & \textcolor{green}{\XSolidBrush}     \\
& EmbMarker & 68.35 $\pm$ 1.32   & $>0.6442$ & 0.72 $\pm$ 5.37  & -1.43 $\pm$ 10.74 & \textcolor{red}{\XSolidBrush}     \\
& WARDEN    & 67.01 $\pm$ 0.18   & $>0.3558$ & 0.00 $\pm$ 4.71   & 0.00 $\pm$ 9.41 & \textcolor{red}{\XSolidBrush}  \\
\rowcolor{gray!20} & EspeW(Ours) & \textbf{65.61 $\pm$ 0.49}   & $\boldsymbol{<10^{-9}}$ & \textbf{32.98 $\pm$ 9.34} & \textbf{-65.96 $\pm$ 18.67} & \textcolor{green}{\Checkmark}  \\
\bottomrule
\end{tabular}
}
\label{tab:main_mind}
\end{table}

\begin{table}[h]
\caption{Performance of different methods on AGNews. For no CSE, lower ACC means better harmlessness. For CSE, lower ACC means better watermark effectiveness. In "COPY?" column, correct verifications are green and failures are red. Best results are highlighted in \textbf{bold} (except Original). }
\centering
\scalebox{0.90}{
\begin{tabular}{@{}ccclrrc@{}}
\toprule
\multicolumn{1}{c}{$K$(CSE)} & \multicolumn{1}{c}{Method}  & \multicolumn{1}{c}{ACC$(\%)\downarrow$} & \multicolumn{1}{c}{$p$-value$\downarrow$} & \multicolumn{1}{c}{$\Delta \cos(\%)\uparrow$} & \multicolumn{1}{c}{$\Delta l_{2}(\%)\downarrow$} & \multicolumn{1}{c}{COPY?} \\ \midrule
\multirow{4}{*}{No CSE}              
                                 & Original  & 93.43 $\pm$ 0.27   & $>$0.02324 & 1.11 $\pm$ 0.42  & -2.22 $\pm$ 0.83    & \textcolor{green}{\XSolidBrush} \\
                                 & EmbMarker & \textbf{93.60 $\pm$ 0.06}   & $\boldsymbol{<10^{-11}}$ & \textbf{13.15 $\pm$ 0.55}   & \textbf{-26.29 $\pm$ 1.11} & \textcolor{green}{\Checkmark}   \\
                                 & WARDEN    & 93.22 $\pm$ 0.10   & $>$0.0083 & -6.24 $\pm$ 5.96   & 12.47 $\pm$ 11.92 & \textcolor{red}{\XSolidBrush}   \\
\rowcolor{gray!20}               & EspeW(Ours)     & 93.42 $\pm$ 0.16   & $\boldsymbol{<10^{-11}}$ & 9.59 $\pm$ 0.74   & -19.19 $\pm$ 1.49 & \textcolor{green}{\Checkmark}   \\ \cmidrule(l){1-7} 
\multirow{4}{*}{1}               
                                 & Original  & 94.12 $\pm$ 0.14   & $>$0.3936 & 2.22 $\pm$ 0.98   & -4.45 $\pm$ 1.96 & \textcolor{green}{\XSolidBrush}    \\
                                 & EmbMarker & 94.01 $\pm$ 0.18   & $\boldsymbol{<10^{-11}}$  & \textbf{136.32 $\pm$ 2.24}  & \textbf{-272.65 $\pm$ 4.48} & \textcolor{green}{\Checkmark}  \\
                                 & WARDEN    & \textbf{93.75 $\pm$ 0.23}   & $\boldsymbol{<10^{-11}}$  & 96.69 $\pm$ 1.62  & -193.38 $\pm$ 3.24 & \textcolor{green}{\Checkmark}  \\
\rowcolor{gray!20}               & EspeW(Ours)     & 94.05 $\pm$ 0.15   & $\boldsymbol{<10^{-11}}$  & 56.51 $\pm$ 2.47  & -113.02 $\pm$ 4.95 & \textcolor{green}{\Checkmark}  \\ \cmidrule(l){1-7} 
\multirow{4}{*}{50}              
                                 & Original  & 93.39 $\pm$ 0.24   & $>$0.0454 & -4.78 $\pm$ 1.03   & 9.56 $\pm$ 2.05 & \textcolor{green}{\XSolidBrush}    \\
                                 & EmbMarker & 93.04 $\pm$ 0.33   & $<10^{-6}$  & 14.43 $\pm$ 4.91  & -28.85 $\pm$ 9.81 & \textcolor{green}{\Checkmark}  \\
                                 & WARDEN    & \textbf{92.54 $\pm$ 0.36}   & $>$0.3062  & 2.40 $\pm$ 2.32  & -4.79 $\pm$ 4.65 & \textcolor{red}{\XSolidBrush}   \\
\rowcolor{gray!20}               & EspeW(Ours)     & 93.00 $\pm$ 0.12   & $\boldsymbol{<10^{-10}}$  & \textbf{21.83 $\pm$ 5.11}  & \textbf{-43.65 $\pm$ 10.22} & \textcolor{green}{\Checkmark}  \\ \cmidrule(l){1-7} 
\multirow{4}{*}{100}             
                                 & Original  & 92.77 $\pm$ 0.28   & $>$0.0520 & -4.50 $\pm$ 0.66   & 9.00 $\pm$ 1.33 & \textcolor{green}{\XSolidBrush}    \\
                                 & EmbMarker & 92.46 $\pm$ 0.17   & $>$0.0206 & 8.36 $\pm$ 3.72  & -16.71 $\pm$ 7.44 & \textcolor{red}{\XSolidBrush}   \\
                                 & WARDEN    & \textbf{91.62 $\pm$ 0.21}   & $>$0.1488 & -3.95 $\pm$ 2.19  & 7.89 $\pm$ 4.37 & \textcolor{red}{\XSolidBrush}    \\
\rowcolor{gray!20}               & EspeW(Ours)     & 92.81 $\pm$ 0.18   & $\boldsymbol{<10^{-5}}$  & \textbf{20.07 $\pm$ 10.23}  & \textbf{-40.15 $\pm$ 20.46} & \textcolor{green}{\Checkmark}  \\ \cmidrule(l){1-7} 
\multirow{4}{*}{1000}            
                                 & Original  & 88.55 $\pm$ 0.21   & $>$0.1745 & 3.4 $\pm$ 0.96  & -6.81 $\pm$ 1.34 & \textcolor{green}{\XSolidBrush}     \\
                                 & EmbMarker & 90.22 $\pm$ 0.31   & $>$0.8320 & 2.58 $\pm$ 2.18   & -5.17 $\pm$ 3.12 & \textcolor{red}{\XSolidBrush}     \\
                                 & WARDEN    & \textbf{79.82 $\pm$ 0.22}   & $>$0.0335 & -6.51 $\pm$ 3.96   & 13.03 $\pm$ 6.76 & \textcolor{red}{\XSolidBrush}  \\
\rowcolor{gray!20}               & EspeW(Ours)      & 86.92 $\pm$ 0.19   & $\boldsymbol{<10^{-8}}$ & \textbf{23.03 $\pm$ 11.12} & \textbf{-46.07 $\pm$ 23.12} & \textcolor{green}{\Checkmark}  \\ \bottomrule
\end{tabular}}
\label{tab:main_agnews}
\end{table}

\begin{table}[h]
\caption{Performance of different methods on Enron Spam. For no CSE, higher ACC means better harmlessness. For CSE, lower ACC means better watermark effectiveness. In "COPY?" column, correct verifications are green and failures are red. Best results are highlighted in \textbf{bold} (except Original). }
\centering
\scalebox{0.90}{
\begin{tabular}{@{}ccclrrc@{}}
\toprule
\multicolumn{1}{c}{$K$(CSE)} & \multicolumn{1}{c}{Method}  & \multicolumn{1}{c}{ACC$(\%)$} & \multicolumn{1}{c}{$p$-value$\downarrow$} & \multicolumn{1}{c}{$\Delta \cos(\%)\uparrow$} & \multicolumn{1}{c}{$\Delta l_{2}(\%)\downarrow$} & \multicolumn{1}{c}{COPY?} \\ \midrule
\multirow{4}{*}{No CSE}              
                                 & Original  & 94.90$\pm$0.35   & $>$0.5776 & -0.11$\pm$0.26  & 0.22$\pm$0.52    & \textcolor{green}{\XSolidBrush} \\
                                 & EmbMarker & \textbf{94.86$\pm$0.24}   & $<10^{-10}$ & \textbf{9.75$\pm$0.11}   & \textbf{-19.49$\pm$0.21} & \textcolor{green}{\Checkmark}   \\
                                 & WARDEN    & 94.31$\pm$0.44   & $\boldsymbol{<10^{-11}}$ & 7.00$\pm$0.62   & -14.00$\pm$1.24 & \textcolor{green}{\Checkmark}   \\
\rowcolor{gray!20} &  EspeW(Ours)     & 94.73$\pm$0.23   & $<10^{-10}$ & 7.23$\pm$0.35   & -14.47$\pm$0.70 & \textcolor{green}{\Checkmark}   \\ \cmidrule(l){1-7} 
\multirow{4}{*}{1}               
                                 & Original  & 95.99$\pm$0.41   & $>$0.5791 & 0.58$\pm$2.06   & -1.15$\pm$4.12 & \textcolor{green}{\XSolidBrush}    \\
                                 & EmbMarker & 95.93$\pm$0.37   & $<10^{-10}$  & \textbf{69.55$\pm$7.16}  & \textbf{-139.10$\pm$14.32} & \textcolor{green}{\Checkmark}  \\
                                 & WARDEN    & \textbf{95.80$\pm$0.05}   & $\boldsymbol{<10^{-11}}$  & 68.01$\pm$1.62  & -136.02$\pm$3.23 & \textcolor{green}{\Checkmark}  \\
\rowcolor{gray!20} &  EspeW(Ours)     & 95.86$\pm$0.19   & $<10^{-10}$  & 56.25$\pm$3.53  & -112.50$\pm$7.06 & \textcolor{green}{\Checkmark}  \\ \cmidrule(l){1-7} 
\multirow{4}{*}{50}              
                                 & Original  & 95.68$\pm$0.13   & $>$0.7668 & 0.50$\pm$1.15   & -1.00$\pm$2.30 & \textcolor{green}{\XSolidBrush}    \\
                                 & EmbMarker & 95.48$\pm$0.47   & $>$0.0002  & 11.00$\pm$1.77  & -22.01$\pm$3.53 & \textcolor{red}{\XSolidBrush}  \\
                                 & WARDEN    & \textbf{95.39$\pm$0.14}   & $>$0.5751  & -1.39$\pm$2.38  & 2.77$\pm$4.77 & \textcolor{red}{\XSolidBrush}   \\
\rowcolor{gray!20} &  EspeW(Ours)     & 95.48$\pm$0.28   & $\boldsymbol{<10^{-10}}$  & \textbf{47.75$\pm$4.13}  & \textbf{-95.50$\pm$8.26} & \textcolor{green}{\Checkmark}  \\ \cmidrule(l){1-7} 
\multirow{4}{*}{100}             
                                 & Original  & 95.44$\pm$0.54   & $>$0.6805 & 0.45$\pm$0.73   & -0.91$\pm$1.46 & \textcolor{green}{\XSolidBrush}    \\
                                 & EmbMarker & 95.34$\pm$0.31   & $>$0.0114 & 10.75$\pm$2.91  & -21.50$\pm$5.82 & \textcolor{red}{\XSolidBrush}   \\
                                 & WARDEN    & \textbf{94.86$\pm$0.29}   & $>$0.4970 & -0.13$\pm$4.28  & 0.25$\pm$8.57 & \textcolor{red}{\XSolidBrush}    \\
\rowcolor{gray!20} &  EspeW(Ours)     & 95.25$\pm$0.30   & $\boldsymbol{<10^{-10}}$  & \textbf{44.24$\pm$6.44}  & \textbf{-88.49$\pm$12.87} & \textcolor{green}{\Checkmark}  \\ \cmidrule(l){1-7} 
\multirow{4}{*}{1000}            
                                 & Original  & 94.69$\pm$0.26   & $>$0.4169 & -1.17$\pm$2.05  & 2.33$\pm$4.10 & \textcolor{green}{\XSolidBrush}     \\
                                 & EmbMarker & 94.89$\pm$0.54   & $>$0.0243 & 6.66$\pm$2.63   & -13.32$\pm$5.26 & \textcolor{red}{\XSolidBrush}     \\
                                 & WARDEN    & \textbf{94.39$\pm$0.41}   & $>$0.3736 & 2.45$\pm$4.32   & -4.91$\pm$8.63 & \textcolor{red}{\XSolidBrush}  \\
\rowcolor{gray!20} &  EspeW(Ours)      & 94.69$\pm$0.66   & $\boldsymbol{<10^{-9}}$ & \textbf{35.25$\pm$3.29} & \textbf{-70.51$\pm$6.58} & \textcolor{green}{\Checkmark}  \\ \bottomrule
\end{tabular}}
\label{tab:main_enron}
\end{table}

\subsection{Ablation Results on More Datasets}
\label{sec:ablation_results_on_more_datasets}

\begin{figure}[h]
    \centering
    \begin{subfigure}[t]{0.48\textwidth}
        \centering
        \includegraphics[width=\textwidth]{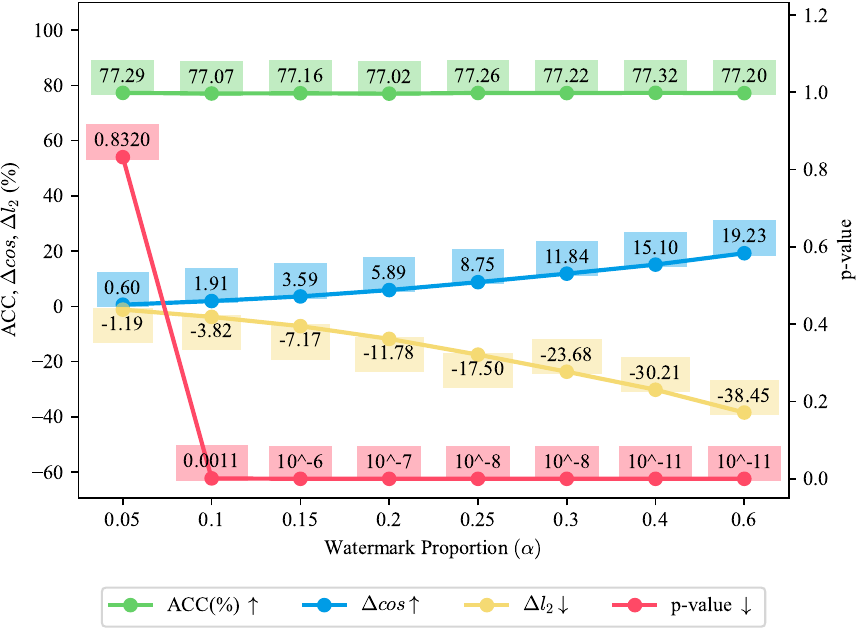}
        \caption{Effect of watermark proportion without CSE.}
        \label{fig:ablation_alpha_no_CSE_mind}
    \end{subfigure}
    \hspace{0.02\textwidth}
    \begin{subfigure}[t]{0.48\textwidth}
        \centering
        \includegraphics[width=\textwidth]{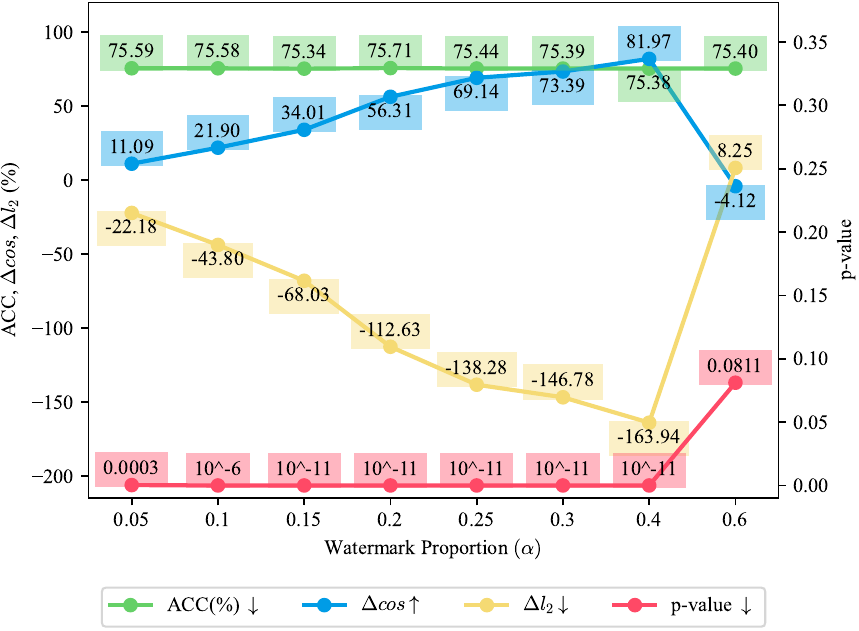}
        \caption{Effect of watermark proportion with CSE.}
        \label{fig:ablation_alpha_CSE_mind}
    \end{subfigure}
    \caption{Ablation results of watermark proportion on MIND. (a) shows results without CSE. (b) shows results with CSE, where $K$ is set to 50. }
    \label{fig:ablation_alpha_mind}
\end{figure}

\begin{figure}[h]
    \centering
    \begin{subfigure}[t]{0.48\textwidth}
        \centering
        \includegraphics[width=\textwidth]{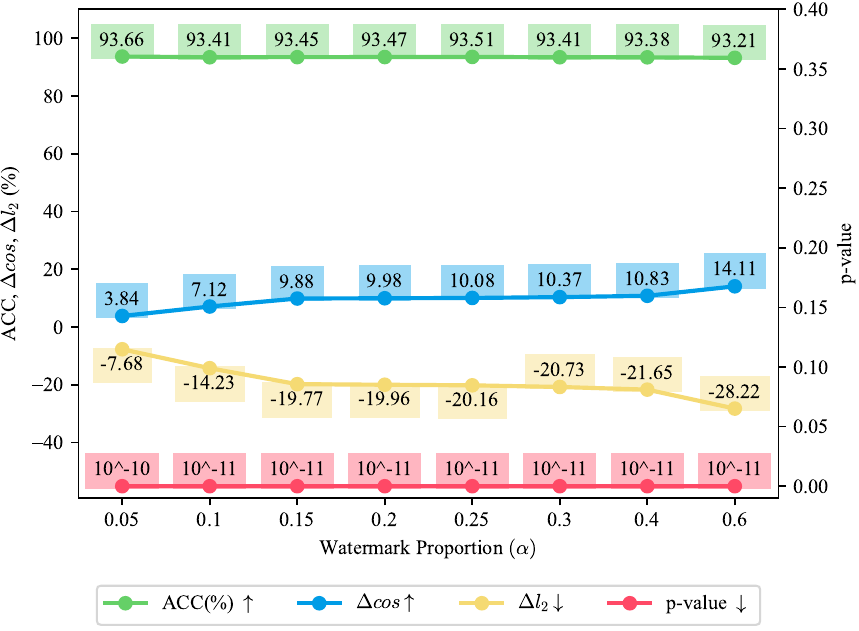}
        \caption{Effect of watermark proportion without CSE.}
        \label{fig:ablation_alpha_no_CSE_agnews}
    \end{subfigure}
    \hspace{0.02\textwidth}
    \begin{subfigure}[t]{0.48\textwidth}
        \centering
        \includegraphics[width=\textwidth]{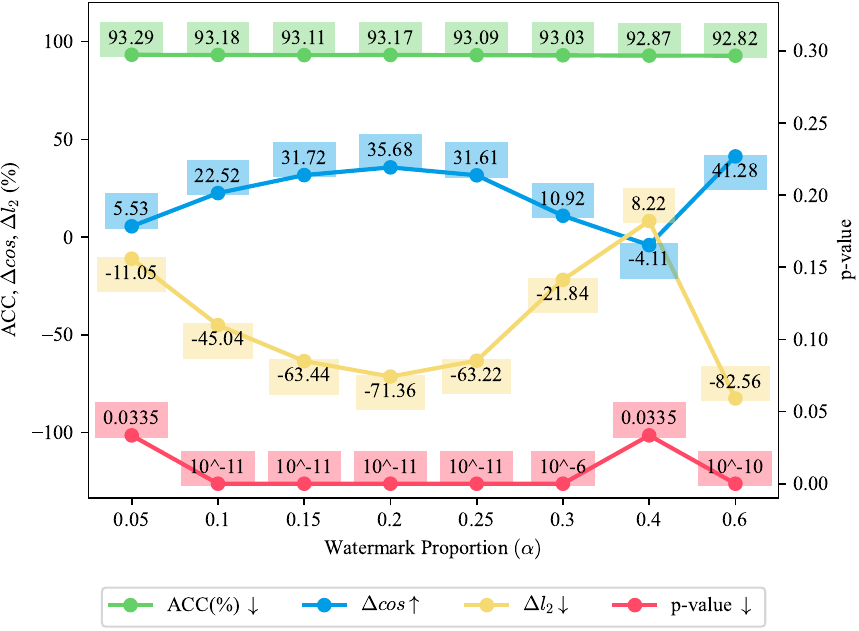}
        \caption{Effect of watermark proportion with CSE.}
        \label{fig:ablation_alpha_CSE_agnews}
    \end{subfigure}
    \caption{Ablation results of watermark proportion on AGNews. (a) shows results without CSE. (b) shows results with CSE, where $K$ is set to 50. }
    \label{fig:ablation_alpha_agnews}
\end{figure}

\begin{figure}[h]
    \centering
    \begin{subfigure}[t]{0.48\textwidth}
        \centering
        \includegraphics[width=\textwidth]{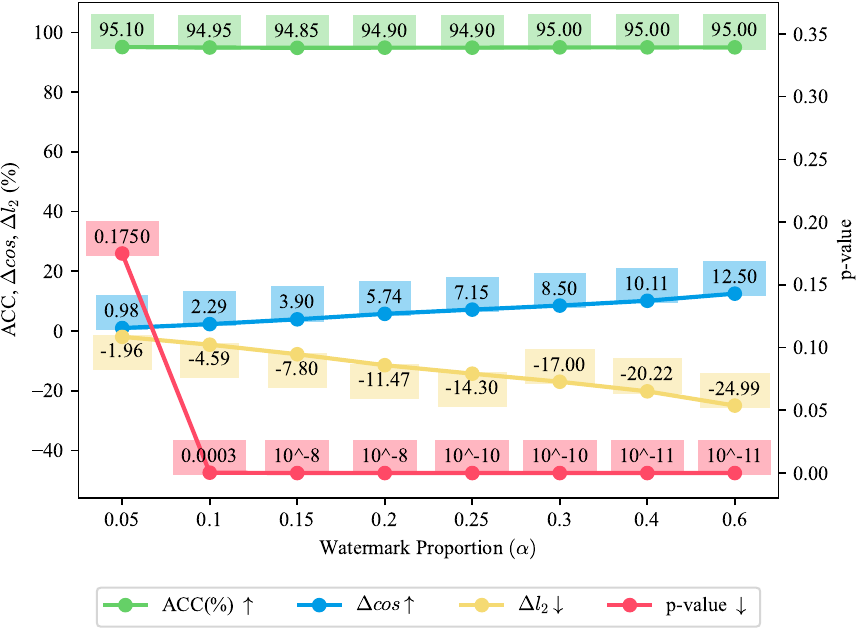}
        \caption{Effect of watermark proportion without CSE.}
        \label{fig:ablation_alpha_no_CSE_enron}
    \end{subfigure}
    \hspace{0.02\textwidth}
    \begin{subfigure}[t]{0.48\textwidth}
        \centering
        \includegraphics[width=\textwidth]{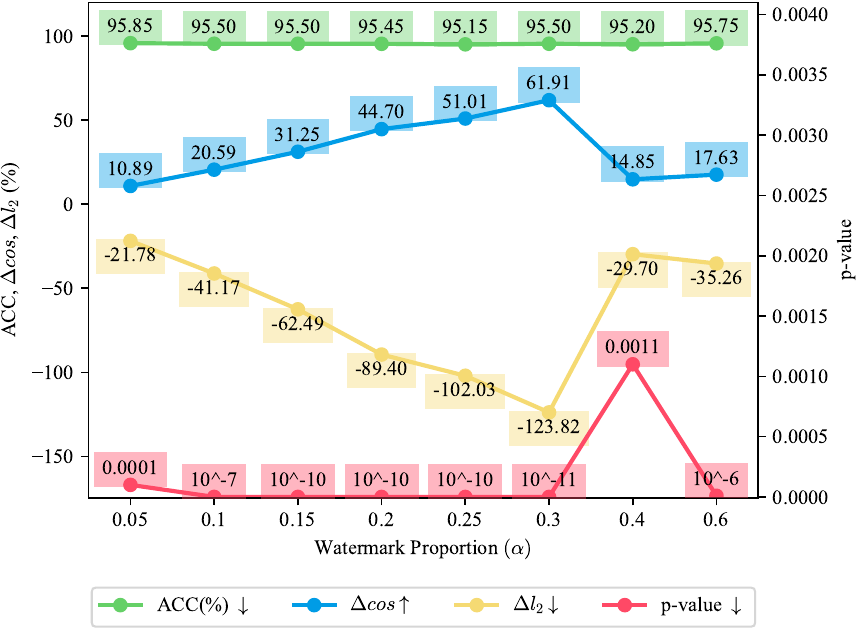}
        \caption{Effect of watermark proportion with CSE.}
        \label{fig:ablation_alpha_CSE_enron}
    \end{subfigure}
    \caption{Ablation results of watermark proportion on Enron Spam. (a) shows results without CSE. (b) shows results with CSE, where $K$ is set to 50. }
    \label{fig:ablation_alpha_enron}
\end{figure}

We present additional ablation results on other datasets in Fig.~\ref{fig:ablation_alpha_mind}, Fig.~\ref{fig:ablation_alpha_agnews}, and Fig.~\ref{fig:ablation_alpha_enron}. 
When CSE is not applied, it can be observed that our proposed method can inject watermark successfully with a minimum $\alpha$ value of 15\% on all datasets. And as $\alpha$ increases, the detection performance of the watermark is also greater. When CSE is applied, compared with the situation without CSE, the trend in detection performance relative to $\alpha$ remains similar when $\alpha$ is small. However, when a large $\alpha$ is set, our method will fail. These findings are consistent with those on the SST2 dataset. 

\subsection{Understanding the Enhanced Detection of ESpeW Under Stronger CSE Attacks}
\label{sec:understanding_cse}

The watermark (WM) detection performance of ESpeW increases under stronger CSE attacks (higher $K$) due to the fundamental mechanism of CSE. Specifically, CSE aims to remove the principal components of the embeddings that are commonly shared among multiple embeddings. These principal components include both watermark-related and watermark-unrelated components. 

Watermark-unrelated components are inherently present in embeddings—even without an intentionally embedded watermark, two embeddings will exhibit a certain degree of similarity. Our method, however, ensures that watermark-related components are resistant to removal. Consequently, as $K$ increases, CSE tends to eliminate a greater proportion of watermark-unrelated components. 

To further illustrate this phenomenon, we calculate the cosine similarities between the target embedding and the embeddings of both clean samples and trigger-bearing samples, as shown in Table~\ref{tab:cse_impact}. Additionally, we present the ratio between these two similarities, which serves as an indicator of watermark detection effectiveness. A higher ratio indicates stronger detection capability. 

\begin{table}[t]
    \centering
    \begin{tabular}{lccc}
        \toprule
        $K$ & \textbf{Cos Sim (Clean, Target)} & \textbf{Cos Sim (Trigger, Target)} & \textbf{Ratio (Trigger/Clean)} \\
        \midrule
        No CSE  & 0.8262 & 0.9405 & 1.14 \\
        1       & 0.0620 & 0.7220 & 11.65 \\
        50      & 0.0539 & 0.6717 & 12.46 \\
        1000    & 0.0226 & 0.4096 & 18.12 \\
        \bottomrule
    \end{tabular}
    \caption{Impact of CSE attack strength ($K$) on cosine similarity and watermark detection.}
    \label{tab:cse_impact}
\end{table}

The results demonstrate (1) Without CSE, a notable similarity exists between the target embedding and clean embeddings. (2) For low-intensity CSE ($K=1$), the similarity between clean embeddings and the target embedding decreases significantly, while the similarity for trigger-bearing embeddings remains relatively high. (3) As the attack intensity increases ($K=50, 1000$), this trend becomes more pronounced. 

This phenomenon is not unique to our method and is also observed in other watermarking techniques such as EmbMarker and WARDEN under $K=1$, where their performance is enhanced compared to the absence of an attack (see Tab.~\ref{tab:main_sst2}).

\subsection{Ablation on Impact of $\alpha$ on Cosine Similarity between Original and Watermarked Embedding.}
\label{sec:ablation_alpha_cos}

We conducted ablation experiments to analyze the impact of the parameter $\alpha$ on the cosine similarity between the original embedding and the watermarked embedding. The results are presented in Tab.~\ref{tab:ablation_cos}. As expected, increasing $\alpha$ leads to a decline in embedding quality while enhancing the detection capability. This observation aligns with our understanding and further validates our approach. 

\begin{table}[h!]
    \centering
    \begin{tabular}{ccccccc}
        \toprule
        $\alpha$ & 0.05 & 0.10 & 0.15 & 0.20 & 0.25 & 0.30 \\
        \midrule
        \textbf{cos sim} & 0.9979 & 0.9958 & 0.9936 & 0.9912 & 0.9886 & 0.9856 \\
        \textbf{p-value} & 0.3356 & 0.0040 & $10^{-6}$ & $10^{-8}$ & $10^{-11}$ & $10^{-11}$ \\
        \bottomrule
    \end{tabular}
    \caption{Ablation study on the impact of $\alpha$ on cosine similarity and detection capability.}
    \label{tab:ablation_cos}
\end{table}

\subsection{Testing on More Base Models}
\label{sec:ablation_more_base_model}

We further extend our evaluation by providing comparative experimental results for both BERT-Base-Cased and BERT-Large-Cased models. The results are summarized in Tab.~\ref{tab:bert_comparison}. As shown, the differences between the two models are minimal, indicating that our method maintains its effectiveness across different base models. These results demonstrate the robustness of our method, showing consistent performance across different model architectures. 

\begin{table}[h!]
    \centering
    \begin{tabular}{lcccccc}
        \toprule
        Base Model       & Method & ACC(\%)$\downarrow$ & p-value$\downarrow$ & $\Delta$cos(\%)$\uparrow$ & $\Delta l_2$(\%)$\downarrow$ & COPY? \\
        \midrule
        BERT-Base-Cased  & ESpeW  & 86.23   & $1.45 \times 10^{-11}$ & 64.53     & -129.06 & \textcolor{green}{\Checkmark} \\
        BERT-Large-Cased & ESpeW  & 86.82   & $1.45 \times 10^{-11}$ & 65.28     & -130.57 & \textcolor{green}{\Checkmark} \\
        \bottomrule
    \end{tabular}
    \caption{Comparative results on BERT-Base-Cased and BERT-Large-Cased models using ESpeW.}
    \label{tab:bert_comparison}
\end{table}

\subsection{Random Selection}
\label{sec:random_selection}

We begin by providing a detailed description of the random selection algorithm. A direct random selection approach is suboptimal, as the watermarked positions for the same sentence may vary across different queries. An attacker could exploit this variability by making multiple queries to detect or remove the watermark. To mitigate this issue, we propose using the hash value of the embedding (we adopt SHA-256 as our hash function) as a seed, ensuring consistent and repeatable position selection. The algorithm is outlined in Alg.~\ref{alg:random_selection}.

\begin{algorithm}[ht]
\caption{Random Selection Algorithm}
\begin{algorithmic}[1]
\State \textbf{Input:} Original embedding $e_o$, a hash function $HASH(\cdot)$, watermark proportion $\alpha$
\State \textbf{Output:} Selected watermark positions $M$
\State Convert the original embedding $e_o$ into byte format $B_{e_o}$.
\State Generate a random seed $R = HASH(B_{e_o})$.
\State Using $R$ as seed, select $\alpha |e_o|$ random indices $M$.
\State Return $M$. 
\end{algorithmic}
\label{alg:random_selection}
\end{algorithm}

\textbf{We then conduct formalized time complexity analysis.} For smallest-magnitude selection. Using heap sort for the top-k problem is the most common approach, achieving a time complexity of $O(N \log k)$. \textbf{Thus, the total time complexity of smallest-magnitude selection is $O(N \log k)$.} For random selection. Converting $e_o$ to byte format requires $O(N)$, SHA-256 hashing also takes $O(N)$, and selecting random indices needs $O(k)$. \textbf{Therefore, the total time complexity of random selection is $O(2N + k)$.} Considering the high-dimensional nature of embeddings, random selection typically has a much lower time complexity than smallest-magnitude selection.

\textbf{To evaluate the time consumption}, we conduct experiments using two widely-used open-sourced embedding models: NV-Embed-v2~\citep{lee2024nv}, which ranks first in the METB leaderboard~\citep{muennighoff-etal-2023-mteb}, and Stella~\citep{Stella2024}, which is ranked first in the METB leaderboard under 1.5B. We measure the time for 2,000 generations, repeating the experiment five times to reduce the impact of random fluctuations. All experiments are performed on an Ubuntu 18.04 system with an AMD EPYC 7Y83 64-core CPU and a NVIDIA RTX 4090 GPU. The results are summarized in Tab.~\ref{tab:time_comparison}. As can be seen, the time consumed by the random selection-based watermark is significantly smaller than the model inference time, and it also shows a clear advantage over the smallest-magnitude selection-based watermark in terms of time consumption. Therefore, when scaling to large-scale usage, the random selection method offers a clear advantage. 

\begin{table}[ht]
\centering
\caption{Time consumption comparison between random and smallest-magnitude selection.}
\renewcommand{\arraystretch}{1.2}
\scalebox{0.9}{
\begin{tabular}{cccccc}
\hline
\textbf{Model}       & \textbf{\begin{tabular}[c]{@{}c@{}}Model\\ Size\end{tabular}} & \textbf{\begin{tabular}[c]{@{}c@{}}Embedding \\ Size\end{tabular}} & \textbf{\begin{tabular}[c]{@{}c@{}}Inference \\ Time (ms)\end{tabular}} & \textbf{\begin{tabular}[c]{@{}c@{}}Smallest-magnitude \\ Selection Time (ms)\end{tabular}} & \textbf{\begin{tabular}[c]{@{}c@{}}Random Selec- \\ -tion Time (ms)\end{tabular}} \\ \hline
\textbf{Stella}      & 1.5B               & 1024                   & 4371.80 ± 204.80            & 716.30 ± 1.50                              & 31.49 ± 0.40                       \\ \hline
\textbf{NV-Embed-v2} & 7B                  & 4096                   & 13799.46 ± 459.30           & 3761.18 ± 276.59                           & 86.33 ± 0.49                       \\ \hline
\end{tabular}}
\label{tab:time_comparison}
\end{table}

\textbf{Watermark performance comparison of using smallest-magnitude and random Selection.} We report the detection capability, as well as cosine similarity with clean embedding $\cos(\%) \text{w/clean}$ to assess embedding quality. The parameter $K=50$ is used. From the results, we can see that while random selection sacrifices more embedding quality, it achieves better watermarking performance. 

\begin{table}[ht]
\caption{Watermark performance comparison between smallest-magnitude and random selection.}
\renewcommand{\arraystretch}{1.2}
\centering
\begin{tabular}{cccccc}
\hline
\textbf{Dataset} & \textbf{Method} & \textbf{$p$-value$\downarrow$} & \textbf{$\Delta \cos(\%)\uparrow$} & \textbf{$\Delta l_{2}(\%)\downarrow$} & $\cos(\%) \text{w/clean} \uparrow$ \\
\hline
SST2 & Smallest & $ 10^{-11} $ & 65.11 & -130.23 & 99.19 \\
& Random & $ 10^{-11} $ & 72.81 & -145.62 & 98.87 \\
\hline
MIND & Smallest & $ 10^{-11} $ & 72.14 & -144.28 & 99.23 \\
& Random & $ 10^{-11} $ & 77.27 & -154.55 & 98.69 \\
\hline
AGNews & Smallest & $ 10^{-10} $ & 21.83 & -43.65 & 99.27 \\
& Random & $ 10^{-11} $ & 53.13 & -106.27 & 98.97 \\
\hline
Enron Spam & Smallest & $ 10^{-10} $ & 47.75 & -95.5 & 99.21 \\
& Random & $ 10^{-11} $ & 68.38 & -136.75 & 98.92 \\
\hline
\end{tabular}
\label{tab:watermark_comparison}
\end{table}

The above analyses and experiments demonstrate that both smallest-magnitude selection and random selection have their unique advantages, making them suitable for different application scenarios:

\begin{itemize}
    \item \textbf{Smallest-magnitude selection} significantly benefits embedding quality preservation, with modifications to the clean embeddings under 1\%. This is crucial for real-world applications where organizations aim to improve their rankings on leaderboards while protecting their intellectual property.
    \item \textbf{Random selection}, although sacrificing more embedding quality, offers substantial time savings, making it more suitable for product deployment in large-scale applications.
\end{itemize}

We conclude that both approaches are valuable, and users can choose the appropriate method based on their specific application requirements. 

\subsection{Evaluation on More Embedding Models}
\label{sec:more_models}

We apply our watermark to more models to verify our watermark ESpeW's effectiveness. We select two widely-used open-sourced embedding models: (1) NV-Embed-v2~\citep{lee2024nv}, which ranks first in the METB leaderboard~\citep{muennighoff-etal-2023-mteb} and has an embedding dimension of 4096, and (2) Stella-1.5B-V5~\citep{Stella2024}, which is ranked first in the METB leaderboard under 1.5B. Using the Enron spam dataset and $K=50$, we evaluate watermark performance with different $\alpha$. Based on the results in Tab.~\ref{tab:more_models}, we can see that our method remains effective across these embedding models. And it still demonstrates a high detection capability and robustness to CSE. 

\begin{table}[ht]
\caption{Evaluation of ESpeW on additional embedding models. This evaluation is conducted on Enron Spam under CSE attack with $K=50$.}
\label{tab:more_models}
\centering
\renewcommand{\arraystretch}{1.2}
\begin{tabular}{cccccc}
\toprule
& $\alpha$ & $ACC(\%)$ & $p$-value$\downarrow$ & $\Delta \cos(\%)\uparrow$ & $\Delta l_{2}(\%)\downarrow$ \\ \hline
\multirow{7}{*}{Stella}   & 0.05       & 95.69   & 9.55E-06  & 13.12      & -26.23    \\
                          & 0.1        & 95.81   & 1.13E-08  & 27.02      & -54.04    \\
                          & 0.15       & 95.99   & 1.13E-08  & 36.62      & -73.24    \\
                          & 0.2        & 95.39   & 5.80E-10  & 47.30      & -94.60    \\
                          & 0.25       & 95.99   & 5.80E-10  & 56.77      & -113.54   \\
                          & 0.3        & 95.99   & 5.80E-10  & 62.31      & -124.62   \\
                          & 0.6        & 95.32   & 9.55E-06  & 10.45      & -20.89    \\ \hline
\multirow{7}{*}{NV-Embed} & 0.05       & 96.20   & 2.70E-04  & 9.04       & -18.08    \\
                          & 0.1        & 96.10   & 1.13E-08  & 23.90      & -47.79    \\
                          & 0.15       & 95.70   & 5.80E-10  & 40.56      & -81.13    \\
                          & 0.2        & 95.90   & 1.45E-11  & 52.08      & -104.17   \\
                          & 0.25       & 96.25   & 1.45E-11  & 65.99      & -131.98   \\
                          & 0.3        & 95.95   & 1.45E-11  & 72.47      & -144.93   \\
                          & 0.6        & 96.10   & 1.45E-11  & 53.36      & -106.72  \\ \bottomrule
\end{tabular}
\end{table}

\subsection{Resistance Against More Potential Removal Attacks}
\label{sec:more_attack}

\subsubsection{Resistance Against Fine-tuning}
\label{sec:more_attack_finetune}

To evaluate the robustness of our method against fine-tuning attacks, we adopt the unsupervised fine-tuning approach SimCSE~\citep{gao2021simcse}. SimCSE applies contrastive learning by introducing random dropout masks in the Transformer encoder. Positive samples are generated by feeding the same input twice with different dropout masks, while negative samples are constructed from other sentences within the batch. Note that supervised fine-tuning is fundamentally incompatible with embedding models, as it would cause the embeddings to carry excessive label information, compromising semantic properties. Thus, we focus on unsupervised fine-tuning. The experiments are conducted using the hyperparameter settings provided in our paper, and evaluated on the Enron Spam dataset. Fine-tuning parameters are consistent with SimCSE~\citep{gao2021simcse}, using a learning rate of $3 \times 10^{-5}$ and a batch size of 64.

During the detection phase, we replace the p-value with $\Delta \text{cos} (\%)$ and $\Delta l_{2} (\%)$ as evaluation metrics. This adjustment is necessary because fine-tuning induces increased instability in embeddings, causing the p-value to inflate abnormally and lose reliability. To address this, we use the alternative metrics introduced in our paper, ensuring that the false positive rate (FPR) remains below $10^{-5}$ by adjusting the detection thresholds. 

Tab.~\ref{tab:ft_results} demonstrates that our approach effectively defends against fine-tuning attacks, even after 100 epochs of fine-tuning. Considering that data stealing typically involves fewer than 10 epochs, the cost of fine-tuning is significant in practice.

\begin{table}[ht]
\centering
\caption{Performance of our method under SimCSE-based unsupervised fine-tuning attacks. }
\label{tab:ft_results}
\scalebox{0.79}{
\begin{tabular}{ccccccccc}
\hline
\textbf{Epoch} & \textbf{p-value} & $\Delta \text{cos} (\%)$ & $\Delta l_{2} (\%)$ & \textbf{FPR@0.05} & \textbf{FPR@0.01} & \textbf{FPR@$10^{-3}$} & \textbf{FPR@$10^{-4}$} & \textbf{FPR@$10^{-5}$} \\ \hline
0 & 5.8e-10 & 8.10 & -16.21 & \textcolor{green}{\Checkmark} & \textcolor{green}{\Checkmark} & \textcolor{green}{\Checkmark} & \textcolor{green}{\Checkmark} & \textcolor{green}{\Checkmark} \\
1 & 1.1e-8  & 18.45 & -36.91 & \textcolor{green}{\Checkmark} & \textcolor{green}{\Checkmark} & \textcolor{green}{\Checkmark} & \textcolor{green}{\Checkmark} & \textcolor{green}{\Checkmark} \\
2 & 1.4e-7  & 11.92 & -23.84 & \textcolor{green}{\Checkmark} & \textcolor{green}{\Checkmark} & \textcolor{green}{\Checkmark} & \textcolor{green}{\Checkmark} & \textcolor{green}{\Checkmark} \\
3 & 1.3e-6  & 9.11  & -18.23 & \textcolor{green}{\Checkmark} & \textcolor{green}{\Checkmark} & \textcolor{green}{\Checkmark} & \textcolor{green}{\Checkmark} & \textcolor{green}{\Checkmark} \\
4 & 1.4e-7  & 12.42 & -24.83 & \textcolor{green}{\Checkmark} & \textcolor{green}{\Checkmark} & \textcolor{green}{\Checkmark} & \textcolor{green}{\Checkmark} & \textcolor{green}{\Checkmark} \\
5 & 1.1e-3  & 7.91  & -15.81 & \textcolor{green}{\Checkmark} & \textcolor{green}{\Checkmark} & \textcolor{green}{\Checkmark} & \textcolor{green}{\Checkmark} & \textcolor{green}{\Checkmark} \\
6 & 1.1e-8  & 14.12 & -28.24 & \textcolor{green}{\Checkmark} & \textcolor{green}{\Checkmark} & \textcolor{green}{\Checkmark} & \textcolor{green}{\Checkmark} & \textcolor{green}{\Checkmark} \\
7 & 1.3e-6  & 12.33 & -24.66 & \textcolor{green}{\Checkmark} & \textcolor{green}{\Checkmark} & \textcolor{green}{\Checkmark} & \textcolor{green}{\Checkmark} & \textcolor{green}{\Checkmark} \\
8 & 4.0e-3  & 6.56  & -13.12 & \textcolor{green}{\Checkmark} & \textcolor{green}{\Checkmark} & \textcolor{green}{\Checkmark} & \textcolor{green}{\Checkmark} & \textcolor{green}{\Checkmark} \\
9 & 4.0e-3  & 4.39  & -8.77  & \textcolor{green}{\Checkmark} & \textcolor{green}{\Checkmark} & \textcolor{green}{\Checkmark} & \textcolor{green}{\Checkmark} & \textcolor{green}{\Checkmark} \\
10 & 2.7e-4 & 6.21  & -12.42 & \textcolor{green}{\Checkmark} & \textcolor{green}{\Checkmark} & \textcolor{green}{\Checkmark} & \textcolor{green}{\Checkmark} & \textcolor{green}{\Checkmark} \\
20 & 2.7e-4 & 6.80  & -13.60 & \textcolor{green}{\Checkmark} & \textcolor{green}{\Checkmark} & \textcolor{green}{\Checkmark} & \textcolor{green}{\Checkmark} & \textcolor{green}{\Checkmark} \\
35 & 0.03   & 5.82  & -11.64 & \textcolor{green}{\Checkmark} & \textcolor{green}{\Checkmark} & \textcolor{green}{\Checkmark} & \textcolor{green}{\Checkmark} & \textcolor{green}{\Checkmark} \\
50 & 0.08   & 2.21  & -4.42  & \textcolor{green}{\Checkmark} & \textcolor{green}{\Checkmark} & \textcolor{green}{\Checkmark} & \textcolor{green}{\Checkmark} & \textcolor{green}{\Checkmark} \\
100 & 0.34  & 3.60  & -7.19  & \textcolor{green}{\Checkmark} & \textcolor{green}{\Checkmark} & \textcolor{green}{\Checkmark} & \textcolor{green}{\Checkmark} & \textcolor{green}{\Checkmark} \\ \hline
\end{tabular}
}
\end{table}

\begin{table}[ht]
\centering
\caption{Thresholds used for detection metrics to achieve target FPR levels. Validated through 100,000 experiments on non-watermarked models.}
\label{tab:thresholds}
\begin{tabular}{ccc}
\hline
\textbf{FPR} & \textbf{Threshold of $\Delta \text{cos} (\%)$} & \textbf{Threshold of $\Delta l_{2} (\%)$} \\ \hline
0.05      & 0.41 & -1.57 \\
0.01      & 0.59 & -2.32 \\
$10^{-3}$ & 0.82 & -3.16 \\
$10^{-4}$ & 1.08 & -3.93 \\
$10^{-5}$ & 1.09 & -4.10 \\ \hline
\end{tabular}
\end{table}

\subsubsection{Resistance Against Quantization}
\label{sec:more_attack_quantization}

Here, we test the robustness of the watermark against quantization attacks. Specifically, we apply 4-bit quantization to the embeddings. The results, shown in Tab.~\ref{tab:quantization}, indicate that the quantization attack has little to no impact on the watermark's performance. 

\begin{table}[ht]
\renewcommand{\arraystretch}{1.2}
\centering
\begin{tabular}{cccccc}
\hline
\textbf{Dataset} & \textbf{$p$-value$\downarrow$} \\
\hline
SST2  & $ 10^{-11} $ \\ \hline
MIND  & $ 10^{-11} $ \\ \hline
AGNews  & $ 10^{-10} $ \\ \hline
Enron Spam  & $ 10^{-10} $ \\ \hline
\end{tabular}
\caption{Watermark performance under 4-bit quantization when K = 50 in  CSE. }
\label{tab:quantization}
\end{table}

\subsubsection{Resistance Against Paraphrasing Attack}
\label{sec:more_attack_paraphrasing_attack}

We note the recent work proposing paraphrasing attack~\cite{shetty2024wet}. Since the authors did not release their paraphrased texts or the corresponding embeddings, we reproduce their approach to generate embeddings. Specifically, we generated three paraphrased versions for each sample from the SST-2 dataset (containing 67K samples) using LLaMA-3.1-8B-Instruct~\cite{grattafiori2024llama}. Then, we obtained the embeddings of these paraphrased texts using NV-Embed-2~\cite{lee2024nv}, which is one of the most advanced embedding models. 

\noindent To maintain the embedding quality after the paraphrasing attack, we set a threshold $T$. We retain only those paraphrased text embeddings whose cosine similarity with the original text's embedding exceeded $T$. If none of the three paraphrased texts met this criterion, the embeddings were discarded. This filtering is reasonable, as a successful attack should preserve the embedding quality. Extracting low-quality embeddings would be meaningless and would not constitute a valid attack. Finally, we calculate the average of the remaining embeddings to train the stealer's model. To ensure accuracy, we employ multiple random seeds during the experiment.

\noindent The results, presented in Tab.~\ref{tab:results_paraphrasing_attack}, demonstrate the robustness of our method. In other words, the paraphrasing attack is not sufficient to remove our watermark without significantly degrading the embedding quality.

\begin{table}[t]
    \centering
    \begin{tabular}{ccccccc}
        \toprule
        $T$ & Random Seed & ACC(\%)$\downarrow$ & p-value$\downarrow$ & $\Delta$cos(\%)$\uparrow$ & $\Delta l_2$(\%)$\downarrow$ & COPY? \\
        \midrule
        95\% & 0 & 95.07 & $5.80 \times 10^{-10}$ & 9.42 & -18.84 & \textcolor{green}{\Checkmark} \\
        95\% & 1 & 94.84 & $5.80 \times 10^{-10}$ & 10.54 & -21.08 & \textcolor{green}{\Checkmark} \\
        95\% & 2 & 94.84 & $5.80 \times 10^{-10}$ & 10.16 & -20.31 & \textcolor{green}{\Checkmark} \\
        95\% & 3 & 94.95 & $5.80 \times 10^{-10}$ & 10.65 & -21.31 & \textcolor{green}{\Checkmark} \\
        95\% & 4 & 94.84 & $5.80 \times 10^{-10}$ & 11.27 & -22.54 & \textcolor{green}{\Checkmark} \\ \hline
        90\% & 0 & 95.07 & $5.80 \times 10^{-10}$ & 9.63 & -19.26 & \textcolor{green}{\Checkmark} \\
        90\% & 1 & 94.61 & $5.80 \times 10^{-10}$ & 11.05 & -22.10 & \textcolor{green}{\Checkmark} \\
        90\% & 2 & 95.07 & $5.80 \times 10^{-10}$ & 10.62 & -21.24 & \textcolor{green}{\Checkmark} \\
        90\% & 3 & 95.53 & $5.80 \times 10^{-10}$ & 11.51 & -23.02 & \textcolor{green}{\Checkmark} \\
        90\% & 4 & 94.95 & $5.80 \times 10^{-10}$ & 11.14 & -22.28 & \textcolor{green}{\Checkmark} \\
        \bottomrule
    \end{tabular}
    \caption{Effectiveness of the paraphrasing attack on the ESpeW watermark.}
    \label{tab:results_paraphrasing_attack}
\end{table}

\subsubsection{Resistance Against an Adaptive Attack}
\label{sec:more_attack_adaptive_attack}

By statistically analyzing the frequency of values at each position, $e_t$ might be estimated. Based on this motivation, we discuss an adaptive attack based on statistical analysis, named statistical analysis attack (SAA). The algorithm of SAA in shown in Alg~\ref{alg:SAA}. 

\begin{algorithm}[ht]
\caption{Statistical Analysis Attack (SAA)}
\label{alg:SAA}
\begin{algorithmic}[1]
\State \textbf{Input:} Training embedding set of the stealer $DE_{c} \in \mathbb{R}^{N \times M}$, tolerance level $T$, number of neighboring partitions $N_T$
\State \textbf{Output:} Normalized embedding set after attack

\For{each embedding index $i$}
    \State Obtain the embedding array $DE_{c_{i}} \in \mathbb{R}^N$ for index $i$
    \State Partition $DE_{c_{i}}$ into small intervals using $T$ as the step size
    \State Count the number of elements in each partition
    \State Initialize an empty set $SE = \{\}$
    \State Add the partition with the highest number of elements to $SE$
    \If{a partition with a high concentration of elements is identified}
        \State Add this partition and its $N_T$ neighboring partitions to $SE$
    \EndIf
    \State Calculate the upper and lower bounds of $SE$
    \State Set the numbers within this interval to 0
\EndFor
\State Normalize the resulting embedding
\end{algorithmic}
\end{algorithm}


Through this algorithm, we can identify abnormally clustered values, thereby executing the statistical analysis attack. In our experiments, we fix $T$ to a small value of $10^{-4}$ and evaluate the attack performance with varying values of $N_T$. Since the SAA operation negatively affects embedding quality, we measure watermark quality using the cosine similarity between the embedding and the clean embedding, referred to as \textit{cos-clean}. The other parameters remain the same as those in main experiments. 

The results are summarized in Tab.~\ref{tab:saa_results} and demonstrate that this attack cannot successfully remove the watermark without severely damaging the embedding quality. In detail, with $N_T$ set to 200, the p-value based detection becomes ineffective for watermark detection, while the watermark quality degrades to 64.78\% of its original level. When $N_T$ is increased further, to 300 or beyond, the watermark embedding quality continues to degrade, with the \textit{cos-clean} value reaching as low as 45.11\% at $N_T = 300$. 

\begin{table}[ht]
\centering
\renewcommand{\arraystretch}{1.1}
\caption{Performance under Statistical Analysis Attack (SAA) for varying $N_T$. The watermark quality is evaluated using \textit{cos-clean}. Watermark detection performance is evaluted by p-value, $\Delta \text{cos}$, and $\Delta l_2$ are reported.}
\label{tab:saa_results}
\begin{tabular}{ccccc}
\hline
\textbf{$N_T$} & \textbf{p-value}$\downarrow$ & $\Delta \text{cos} (\%)\uparrow$ & $\Delta l_2 (\%)\downarrow$ & \textbf{cos-clean}$\uparrow$ \\ \hline
1  & $5.80 \times 10^{-10}$   & 7.85  & -15.69  & 0.9887 \\
5  & $5.80 \times 10^{-10}$   & 7.84  & -15.69  & 0.9815 \\
10 & $5.80 \times 10^{-10}$   & 7.36  & -14.71  & 0.9738 \\
20 & $5.80 \times 10^{-10}$   & 6.00  & -11.99  & 0.9576 \\
30 & $1.13 \times 10^{-10}$   & 5.67  & -11.34  & 0.9419 \\
100 & $5.80 \times 10^{-10}$  & 7.95  & -15.91  & 0.8276 \\
200 & $0.0011$    & 7.36  & -14.73  & 0.6478 \\
250 & $0.0335$    & 5.24  & -10.48  & 0.5481 \\
300 & $0.0123$    & 2.22  & -4.44   & 0.4511 \\
350 & $0.0123$    & -7.27 & 14.54   & 0.3620 \\
400 & $0.0040$   & -9.99 & 19.98   & 0.2835 \\ \hline
\end{tabular}
\end{table}

\subsection{Embedding Visualization of More Dataset}
\label{sec:Embedding_Visualization_of_More_Dataset}

\begin{figure}[t]
    \centering
    \begin{subfigure}[b]{0.325\textwidth}
        \centering
        \includegraphics[width=1.0\textwidth]{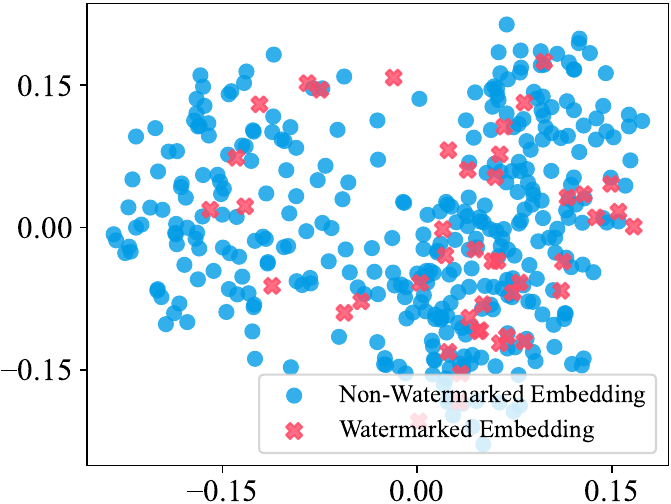}
        \caption{$\alpha = 20\%$}
        \label{fig:embedding_visualization_20_MIND}
    \end{subfigure}
    \begin{subfigure}[b]{0.325\textwidth}
        \centering
        \includegraphics[width=1.0\textwidth]{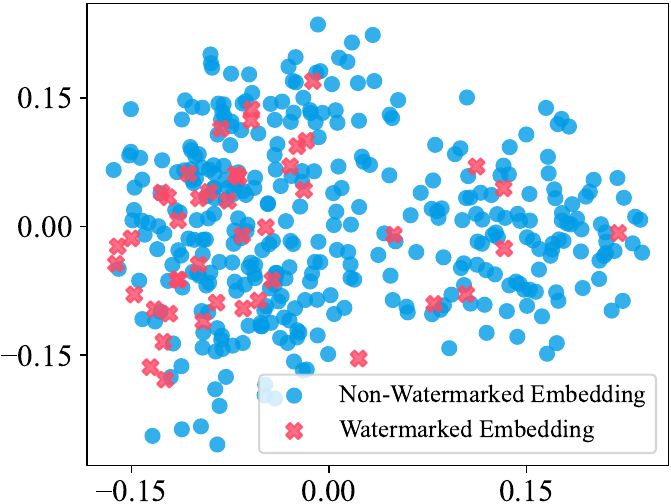}
        \caption{$\alpha = 25\%$}
        \label{fig:embedding_visualization_25_MIND}
    \end{subfigure}
    \begin{subfigure}[b]{0.325\textwidth}
        \centering
        \includegraphics[width=1.0\textwidth]{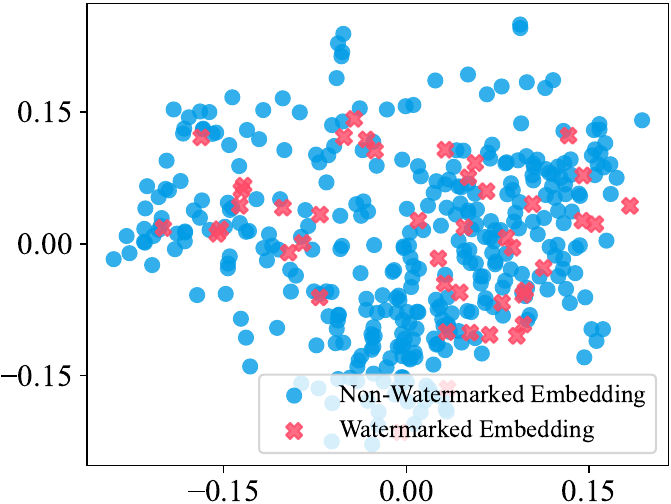}
        \caption{$\alpha = 30\%$}
        \label{fig:embedding_visualization_30_MIND}
    \end{subfigure}
    
    \begin{subfigure}[b]{0.325\textwidth}
        \centering
        \includegraphics[width=1.0\textwidth]{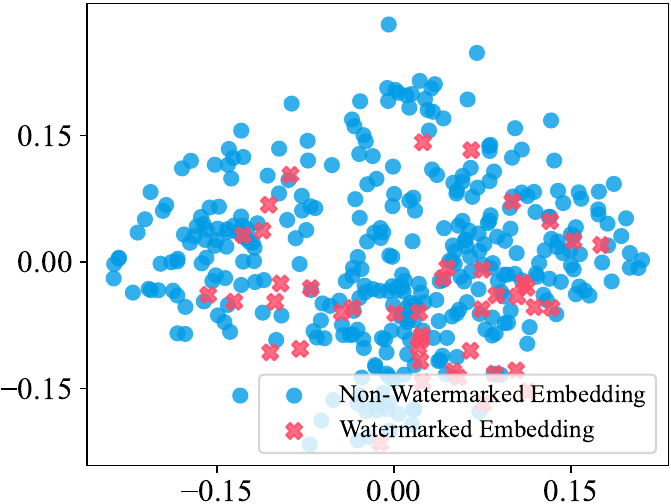}
        \caption{$\alpha = 35\%$}
        \label{fig:embedding_visualization_35_MIND}
    \end{subfigure}
    \begin{subfigure}[b]{0.325\textwidth}
        \centering
        \includegraphics[width=1.0\textwidth]{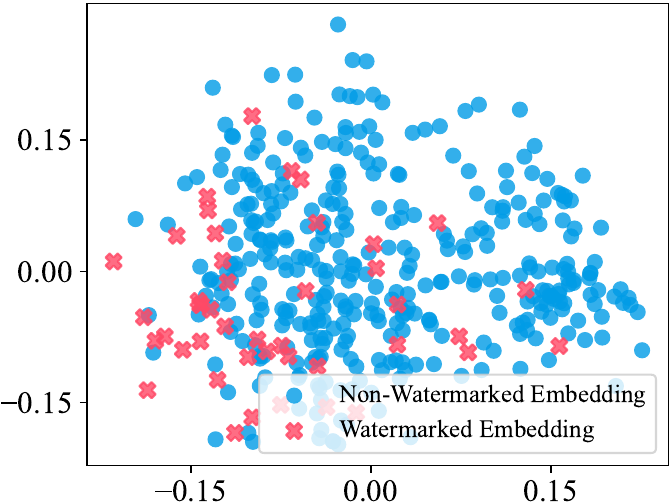}
        \caption{$\alpha = 40\%$}
        \label{fig:embedding_visualization_40_MIND}
    \end{subfigure}
    \begin{subfigure}[b]{0.325\textwidth}
        \centering
        \includegraphics[width=1.0\textwidth]{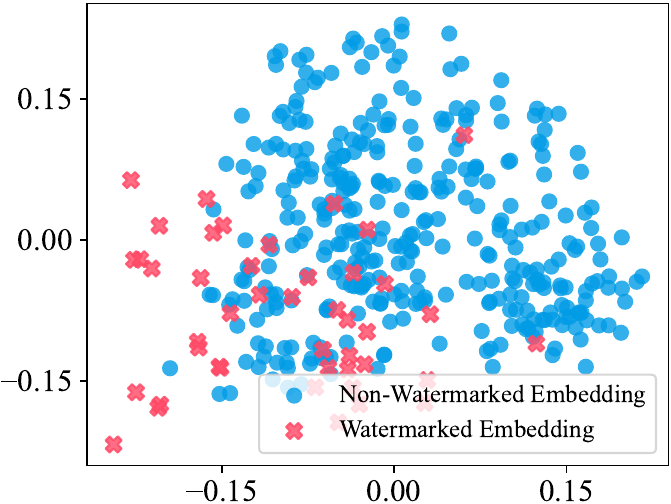}
        \caption{$\alpha = 45\%$}
    \label{fig:embedding_visualization_45_MIND}
    \end{subfigure}
    
    \caption{Visualization of the generated embedding of our ESpeW with different watermark proportion ($\alpha$) on MIND. It shows that we can generate watermarked embeddings indistinguishable with non-watermark embeddings by setting a reasonable watermark proportion. } 
    \label{fig:embedding_visualization_overall_MIND}
\end{figure}

\begin{figure}[t]
    \centering
    \begin{subfigure}[b]{0.325\textwidth}
        \centering
        \includegraphics[width=1.0\textwidth]{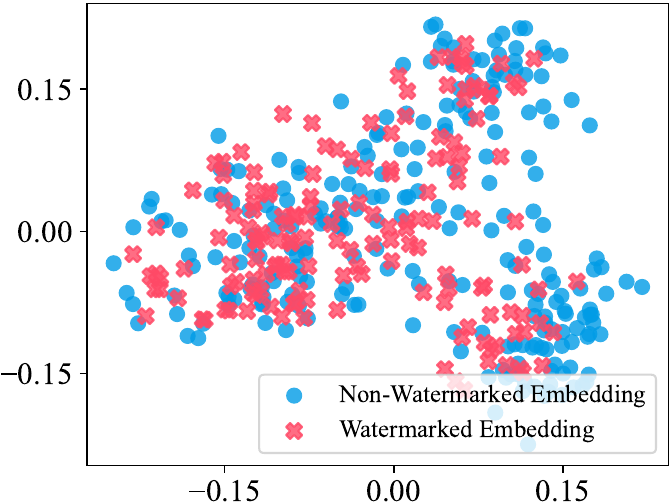}
        \caption{$\alpha = 20\%$}
        \label{fig:embedding_visualization_20_AGNews}
    \end{subfigure}
    \begin{subfigure}[b]{0.325\textwidth}
        \centering
        \includegraphics[width=1.0\textwidth]{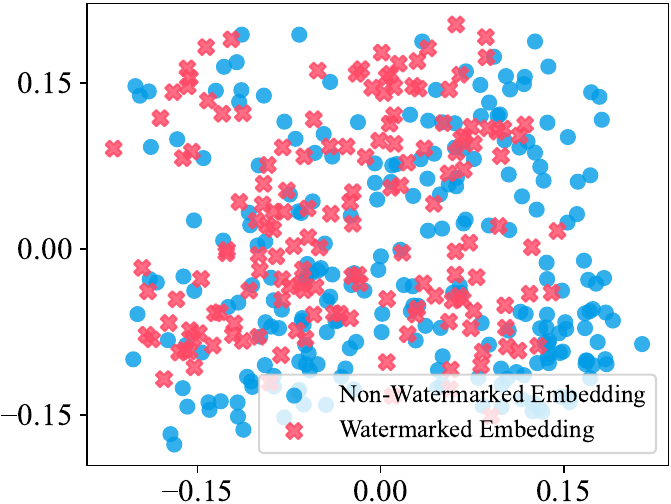}
        \caption{$\alpha = 25\%$}
        \label{fig:embedding_visualization_25_AGNews}
    \end{subfigure}
    \begin{subfigure}[b]{0.325\textwidth}
        \centering
        \includegraphics[width=1.0\textwidth]{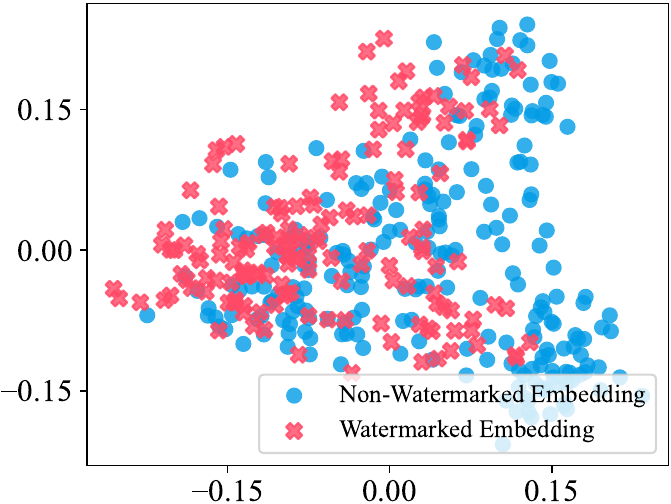}
        \caption{$\alpha = 30\%$}
        \label{fig:embedding_visualization_30_AGNews}
    \end{subfigure}
    
    \begin{subfigure}[b]{0.325\textwidth}
        \centering
        \includegraphics[width=1.0\textwidth]{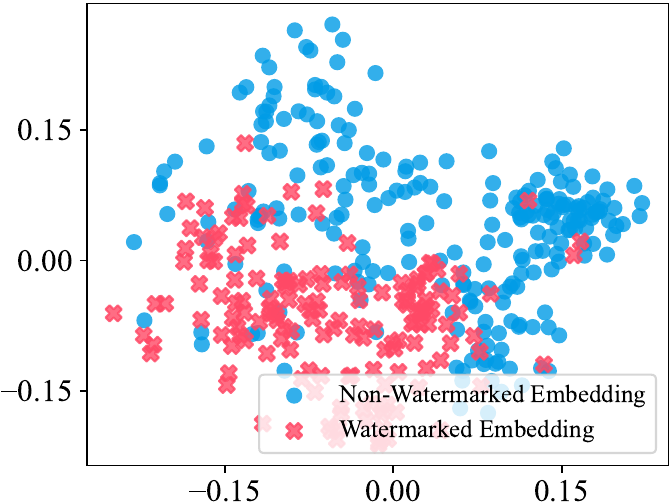}
        \caption{$\alpha = 35\%$}
        \label{fig:embedding_visualization_35_AGNews}
    \end{subfigure}
    \begin{subfigure}[b]{0.325\textwidth}
        \centering
        \includegraphics[width=1.0\textwidth]{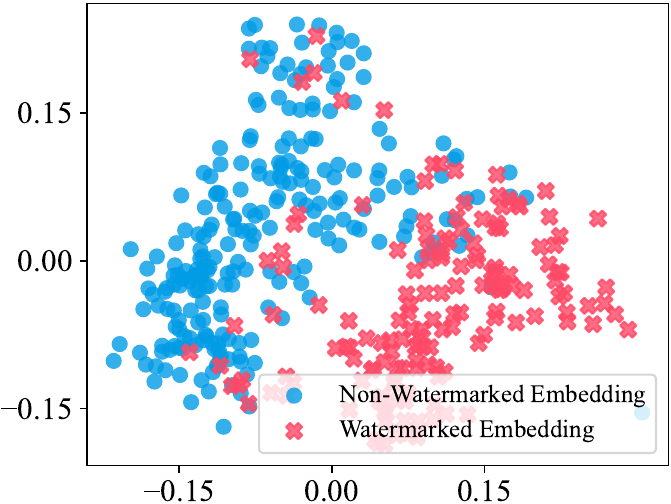}
        \caption{$\alpha = 40\%$}
        \label{fig:embedding_visualization_40_AGNews}
    \end{subfigure}
    \begin{subfigure}[b]{0.325\textwidth}
        \centering
        \includegraphics[width=1.0\textwidth]{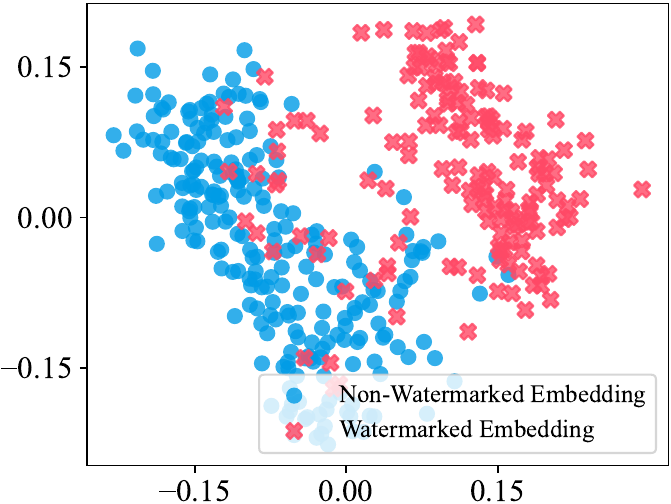}
        \caption{$\alpha = 45\%$}
    \label{fig:embedding_visualization_45_AGNews}
    \end{subfigure}
    
    \caption{Visualization of the generated embedding of our ESpeW with different watermark proportion ($\alpha$) on AGNews. It shows that we can generate watermarked embeddings indistinguishable with non-watermark embeddings by setting a reasonable watermark proportion. } 
    \label{fig:embedding_visualization_overall_AGNews}
\end{figure}

\begin{figure}[t]
    \centering
    \begin{subfigure}[b]{0.325\textwidth}
        \centering
        \includegraphics[width=1.0\textwidth]{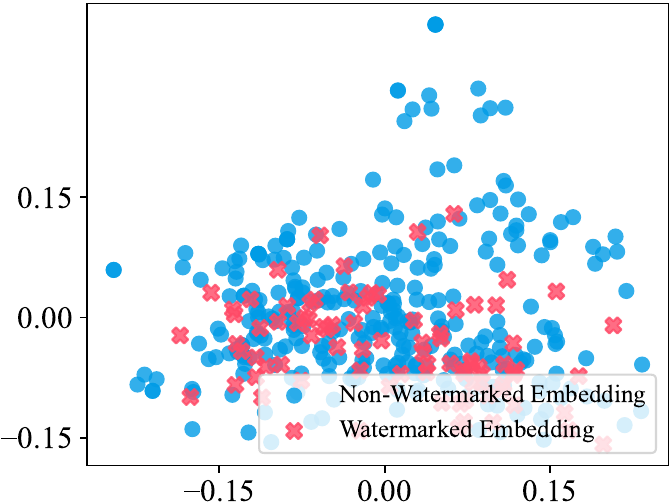}
        \caption{$\alpha = 20\%$}
        \label{fig:embedding_visualization_20_EnronSpam}
    \end{subfigure}
    \begin{subfigure}[b]{0.325\textwidth}
        \centering
        \includegraphics[width=1.0\textwidth]{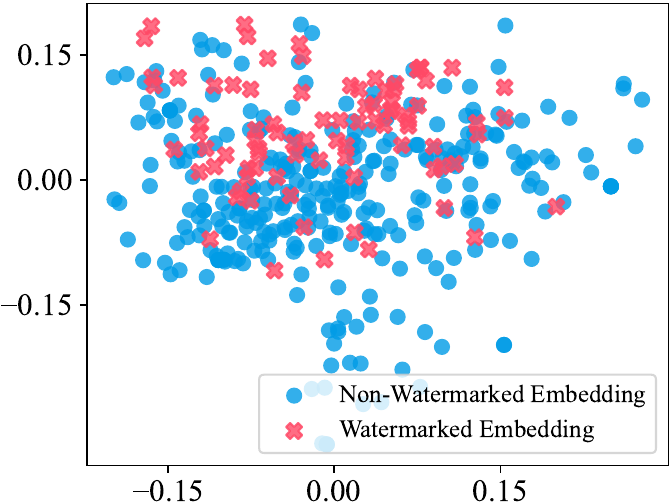}
        \caption{$\alpha = 25\%$}
        \label{fig:embedding_visualization_25_EnronSpam}
    \end{subfigure}
    \begin{subfigure}[b]{0.325\textwidth}
        \centering
        \includegraphics[width=1.0\textwidth]{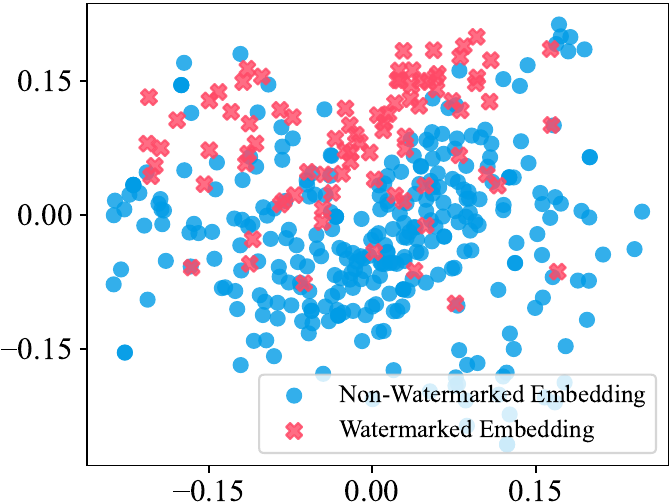}
        \caption{$\alpha = 30\%$}
        \label{fig:embedding_visualization_30_EnronSpam}
    \end{subfigure}
    
    \begin{subfigure}[b]{0.325\textwidth}
        \centering
        \includegraphics[width=1.0\textwidth]{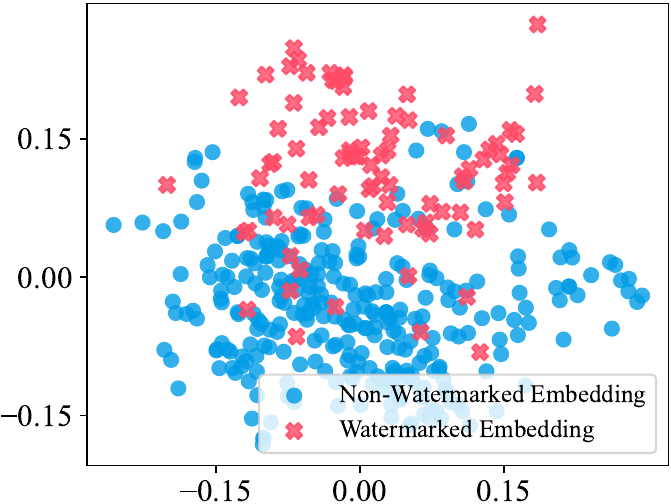}
        \caption{$\alpha = 35\%$}
        \label{fig:embedding_visualization_35_EnronSpam}
    \end{subfigure}
    \begin{subfigure}[b]{0.325\textwidth}
        \centering
        \includegraphics[width=1.0\textwidth]{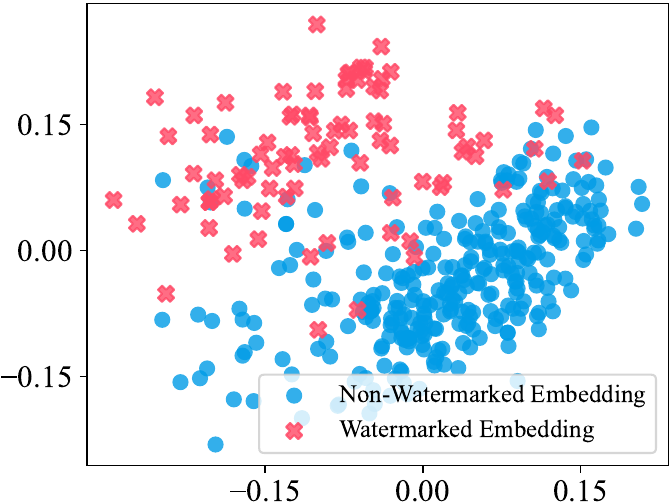}
        \caption{$\alpha = 40\%$}
        \label{fig:embedding_visualization_40_EnronSpam}
    \end{subfigure}
    \begin{subfigure}[b]{0.325\textwidth}
        \centering
        \includegraphics[width=1.0\textwidth]{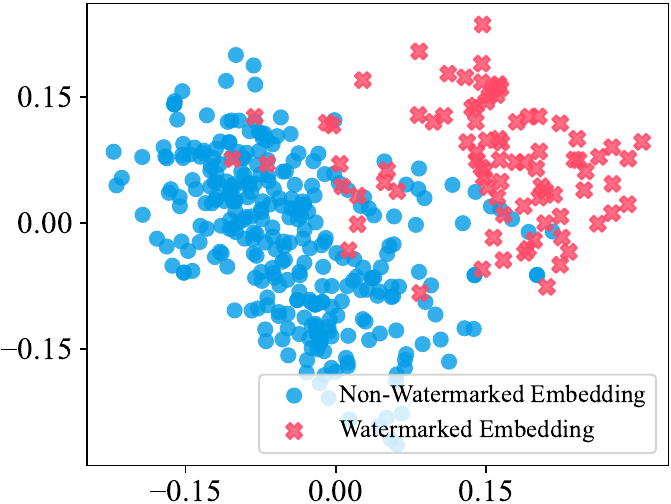}
        \caption{$\alpha = 45\%$}
    \label{fig:embedding_visualization_45_EnronSpam}
    \end{subfigure}
    
    \caption{Visualization of the generated embedding of our ESpeW with different watermark proportion ($\alpha$) on Enron Spam. It shows that we can generate watermarked embeddings indistinguishable with non-watermark embeddings by setting a reasonable watermark proportion. } 
    \label{fig:embedding_visualization_overall_EnronSpam}
\end{figure}

We put more visualization results in Fig.~\ref{fig:embedding_visualization_overall_MIND}, Fig.~\ref{fig:embedding_visualization_overall_AGNews}, and Fig.~\ref{fig:embedding_visualization_overall_EnronSpam}. 

\section{More Discussion}

\subsection{Discussion About Private Key Leakage Scenarios and Corresponding Strategies}
\label{sec:leakage_scenarios}

We here discuss several potential leakage scenarios and corresponding strategies to mitigate these risks. 

\textbf{Leakage Scenarios.} The primary leakage risks are associated with security vulnerabilities, including inadequate storage practices, insecure transmission channels, or insider threats. Inadequate storage, for instance, can result in unauthorized access or accidental exposure of sensitive embeddings. Similarly, insecure transmission of embeddings over unprotected networks can make them vulnerable to interception by malicious actors. Insider threats, where authorized individuals exploit their access for malicious purposes, further exacerbate the risks associated with embedding leakage. These vulnerabilities highlight the need for comprehensive security measures to protect the integrity and confidentiality of target embeddings.

\textbf{Defense Strategies.} To address these risks, we propose several mitigation strategies. One key approach is to regularly renew the security keys used for embedding protection, ensuring that even if a key is compromised, the window of vulnerability is minimized. Additionally, employing multiple keys can help limit the impact of any single breach by compartmentalizing access. It is also crucial to audit and continuously monitor access to sensitive embeddings, enabling quick detection and response to potential security breaches. Encrypting both storage and transmission ensures that even if unauthorized access occurs, the data remains unreadable without the proper decryption keys. Finally, restricting employee access to sensitive information by implementing the principle of least privilege can prevent unnecessary exposure and limit the potential for insider threats.

\subsection{Discussion About False Positive}
\label{sec:false_positive}

Here, we analyze the FPR in our method. In fact, FPR are influenced by most of the parameters discussed in our paper, making it challenging to exhaustively evaluate them under all possible configurations. However, through 100,000 independent tests on non-watermarked models, we can ensure that under the parameter settings used in our paper, the FPR is guaranteed to be less than $10^{-4}$. This represents a remarkably low FPR, which is practical and reliable for real-world applications. 

\subsection{Broader Impacts}
\label{sec:broader_impacts}

Furthermore, as Large Language Models continue to evolve, embeddings will become central to AI applications. However, advanced model theft methods make current service providers reluctant to offer these valuable embeddings. A robust copyright protection method will greatly encourage more service providers to offer embedding services, thereby further accelerating the development and deployment of AI applications. 

\end{document}